\documentclass[journal]{IEEEtran}
\usepackage{amsmath,amsfonts}
\usepackage{array}
\usepackage{graphicx}
\usepackage{cite}
\usepackage{bm}
\usepackage[dvipsnames,table]{xcolor}
\usepackage{booktabs}
\usepackage{multirow}
\usepackage{makecell}
\usepackage{stfloats}
\usepackage{url}
\usepackage{verbatim}
\usepackage{subcaption}
\usepackage{textcomp}
\usepackage{algorithm}
\usepackage{algpseudocode}
\usepackage{tcolorbox}
\usepackage[colorlinks=true,linkcolor=blue,citecolor=blue,urlcolor=blue]{hyperref}
\usepackage[capitalize]{cleveref}
\definecolor{lightblue}{RGB}{220,235,255}
\newcommand{\eg}{e.g.}

\begin{document}

\IEEEpubid{
	\parbox{\textwidth}{\footnotesize\centering
		Copyright \textcopyright\ 2026 IEEE. Personal use of this material is permitted. However, permission to use this material for any other
		purposes must be obtained from the IEEE by sending an email to pubs-permissions@ieee.org.	
		Published in IEEE Transactions on Circuits and Systems for Video Technology. DOI: 10.1109/TCSVT.2026.3745786}
}

\title{SASFormer: Stage-Aware Selective Attention Transformer for All-in-One Blind Image Restoration}

\author{Xiangpeng Tian, Mingyang Li, Fajing Liu, Xiaofeng Wang, Xiaohai He, Chao Ren,~\IEEEmembership{Member,~IEEE}\\
\thanks{This work was supported by the National Natural Science Foundation of China under Grant 62171304, the Science and Technology Project of Xizang Autonomous Region under Grant XZ202601ZY0092, the Young Faculty Technology Innovation Capacity Enhancement Program of Sichuan University under Grant 2024SCUQJTX025, the TCL Science and Technology Innovation Fund under Grant 25JZH008, and the Sichuan Science and Technology Program under Grant 2025JDRC0011. (Corresponding author: Chao Ren.)

Xiangpeng Tian, Chao Ren, Mingyang Li, Fajing Liu, Xiaofeng Wang, and Xiaohai He are with the College of Electronics and Information Engineering, Sichuan University, Chengdu, 610065, China (e-mail: \{tianxp, limingyang, liufajing, xfwang\}@stu.scu.edu.cn, \{hexiaohai, chaoren\}@scu.edu.cn).

Xiangpeng Tian is also with the College of Intelligent Systems Science and Engineering, Hubei Minzu University, Enshi 445000, China.

The source code will be publicly available upon acceptance at https://github.com/TxpHome/SASFormer
}}

\markboth{IEEE Transactions on Circuits and Systems for Video Technology}%
{Tian \MakeLowercase{\textit{et al.}}: SASFormer for All-in-One Blind Image Restoration}

\maketitle

\begin{abstract}

All-in-one image restoration aims to recover high-quality images from multiple degradation types using a single unified model. Recent methods improve unified restoration with hybrid attention, prompt learning, or routing mechanisms, but generally focus on enhancing degradation adaptivity while employing similar feature modeling strategies across network stages. Consequently, the distinct representation requirements of encoding and decoding under diverse degradations remain insufficiently explored.
In this paper, we propose SASFormer, a Stage-Aware Selective Attention Transformer for all-in-one blind image restoration. SASFormer follows a stage-aware design principle: the encoder uses established hybrid channel-spatial attention to build robust content-preserving and degradation-aware representations, while the decoder employs adaptive selective attention for restoration-oriented reconstruction. To address the stage-specific requirements of decoder reconstruction, we develop Adaptive Selective Channel Self-Attention (ASCSA) and Adaptive Selective Spatial Self-Attention (ASSSA), enabling the decoder to dynamically identify restoration-relevant channel responses and spatial contexts. Specifically, ASCSA performs adaptive query-channel selection and attention masking, while ASSSA adaptively integrates spatial attention across multiple window sizes. Compact implicit prompts further modulate the selection process without requiring explicit degradation labels.
Extensive experiments on 3-, 5-, and 11-degradation settings demonstrate that SASFormer achieves strong restoration performance with favorable parameter efficiency. Moreover, stage-wise representation analysis and ablation studies provide empirical support for the proposed stage-aware design and the effectiveness of decoder-side adaptive selection.

\end{abstract}

\begin{IEEEkeywords}
All-in-one blind image restoration, stage-aware representation, adaptive selection, channel-spatial attention, image restoration transformer.
\end{IEEEkeywords}

\section{Introduction}
\IEEEPARstart{I}{mage} restoration aims to reconstruct high-quality images from their degraded counterparts. In recent years, deep learning–based image restoration methods have achieved remarkable progress in handling single-degradation types, such as denoising \cite{Denoise_DeamNet,Denoise_pan2023random,Denoise_lin2023unsupervised}, deraining \cite{Derain_chen2021robust,Derain_li2019heavy}, dehazing \cite{Dehaze_chen2020pmhld,Dehaze_qin2020ffa,Dehaze_shao2020domain}, and deblurring \cite{Deblur_kupyn2019deblurgan,chen2024unsupervised}, significantly improving restoration quality and model performance. These methods mainly explore specific image restoration tasks through various network architectures, including convolutional networks \cite{Denoise_DeamNet,IM_NAFNetchen2022simple,chen2024unsupervised,IM_zamir2021multi}, window-based self-attention \cite{wang2022uformer, liang2021swinir}, channel self-attention \cite{IM_zamir2022restormer}, and hybrid attention mechanisms \cite{chen2024comparative}, achieving remarkable performance. Although these methods have achieved impressive results on specific restoration tasks, their generalization ability remains constrained by the adaptability of the network architecture. Different restoration tasks often require separately trained models, making it challenging to achieve unified multi-task processing without compromising single-task performance, thereby motivating the development of all-in-one restoration frameworks.

\begin{figure*}[tb]
	\centering
	\includegraphics[width=1\linewidth]{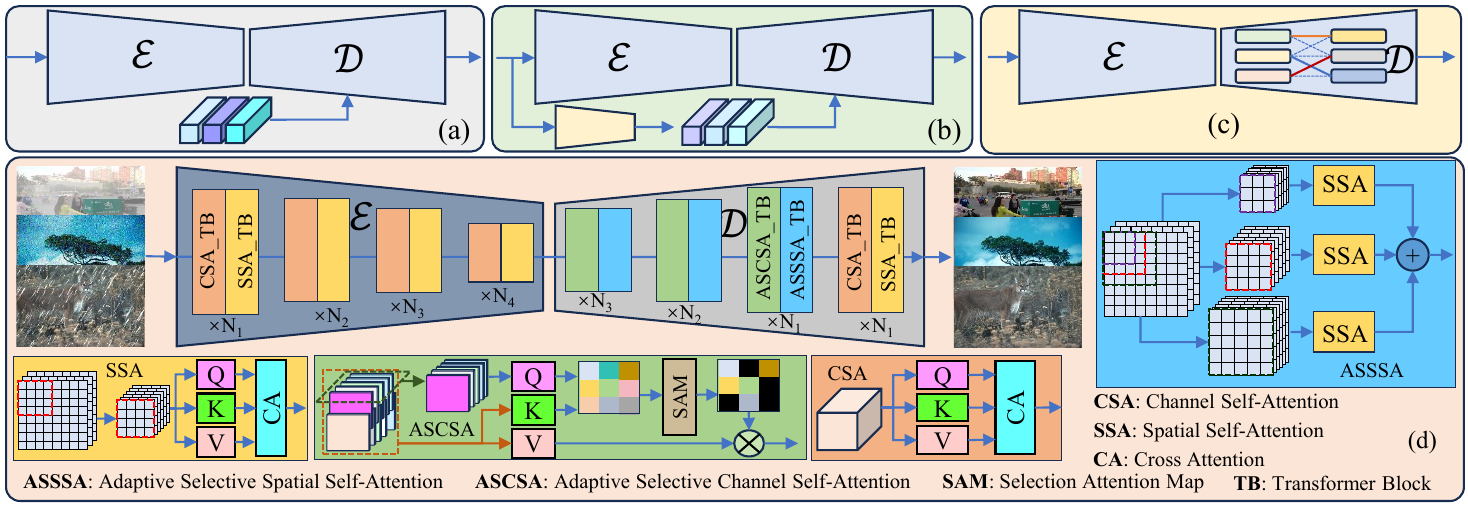} 
	\caption{Existing all-in-one image restoration methods. (a) Adaptive degradation identification methods \cite{potlapalli2023promptir,wu2025learning,cui2025adair}: Utilize degradation priors to guide the restoration process. (b) Auxiliary network or training strategy-based methods \cite{all-in-one_AirNet_li2022all, conde2025instructir, 10970105, Perceive-IR-24, zeng2025vision}: Employ additional networks to identify degradations and guide restoration. (c) Routing or multi-expert methods \cite{zamfir2025complexity, yu2024multi, yang2024all}: Leverage routing strategies or multiple experts for multi-task restoration. (d) Ours: Introduces stage-aware selective attention in the encoder–decoder architecture. The encoder learns robust representations with implicit degradation awareness, while the decoder adaptively selects informative features to suppress degradation interference and improve restoration quality.}
	\label{fig:Introduction-visual}
\end{figure*}

To address this limitation, recent studies have explored all-in-one frameworks that aim to handle diverse image degradations within a unified model~\cite{all-in-one_AirNet_li2022all,potlapalli2023promptir,cui2025adair,10970105}. 
Unlike task-specific models tailored to a single degradation, the key challenge of all-in-one restoration lies in effectively modeling degradation diversity within a shared architecture while maintaining high-quality reconstruction performance.
As illustrated in Fig.~\ref{fig:Introduction-visual}~(a)--(c), existing approaches can be broadly categorized into three groups:
(1) Prompt-based or adaptive degradation recognition methods~\cite{potlapalli2023promptir,wu2025learning,cui2025adair}, which guide the model to identify degradation types through explicit or implicit prompts, or by leveraging frequency-domain differences; 
(2) Auxiliary network or training strategy methods~\cite{all-in-one_AirNet_li2022all, conde2025instructir,10970105,Perceive-IR-24,zeng2025vision}, which introduce additional degradation classification networks or training mechanisms to provide degradation cues; and
(3) Multi-expert or routing-based strategies~\cite{zamfir2025complexity,yu2024multi,yang2024all}, which employ sub-networks or routing mechanisms to optimize performance for different degradations.
From~AirNet \cite{all-in-one_AirNet_li2022all} to PromptIR~\cite{potlapalli2023promptir}, and further to Perceive-IR~\cite{Perceive-IR-24}, VLU-Net~\cite{zeng2025vision}, and~DA-RCOT~\cite{10970105}, research has gradually progressed from simple blind frameworks to more sophisticated degradation modeling approaches. 

\IEEEpubidadjcol

Recent all-in-one image restoration methods have achieved remarkable progress by enhancing degradation awareness. However, existing approaches primarily focus on adapting to diverse degradations, while the distinct representation requirements across different restoration stages remain insufficiently explored. This limitation is particularly relevant because different degradations impose distinct representation demands, while representation learning itself is inherently stage-dependent. Consequently, applying a uniform feature modeling strategy across the entire network may lead to suboptimal representation learning and restoration performance.

To better understand this issue, we revisit the representation requirements of hybrid attention across encoder–decoder architectures and analyze feature behavior across different network stages. We observe that increasing the modeling capacity of hybrid attention does not consistently improve restoration performance. In particular, excessive feature selection in early encoder stages may weaken the preservation of intrinsic image representations, while decoder-stage feature refinement plays a more critical role in reconstructing high-quality image details.
Moreover, feature analysis reveals a clear stage-dependent representation behavior across the network, where encoder representations become progressively more discriminative under different degradations, while decoder representations tend to move toward a more unified representation for clean image reconstruction.

Motivated by this observation, we propose SASFormer, a Stage-Aware Selective Attention Transformer for all-in-one blind image restoration, as shown in Fig.\ref{fig:Introduction-visual}~(d). The key idea is to decouple representation preservation from restoration-oriented feature selection. Rather than applying a uniform attention formulation throughout the network, SASFormer introduces a stage-aware architecture that differentiates feature modeling according to the distinct roles of encoding and decoding. In the encoder, SASFormer adopts established hybrid channel-spatial attention to preserve complementary local and global representations under diverse degradations. In contrast, conventional fixed attention is less suited to decoder reconstruction, where restoration-relevant channel responses and spatial contexts vary across degradations. To address this limitation, the decoder introduces adaptive selective channel and spatial attention to dynamically select informative features, suppress degradation-specific disturbances, and reconstruct intrinsic structures.
Specifically, Adaptive Selective Channel Self-Attention (ASCSA) dynamically selects informative channel subsets and avoids fixed channel partitioning under different degradations. Adaptive Selective Spatial Self-Attention (ASSSA) adaptively combines multiple spatial receptive fields to match degradation-dependent spatial requirements. In addition, implicit prompts are used as compact channel-wise modulation bases for adaptive selection, rather than as spatial feature generators. This design provides lightweight degradation awareness without introducing an auxiliary degradation classifier.

\begin{figure}[tb]
	\centering
	
	\includegraphics[width=1.0\columnwidth]{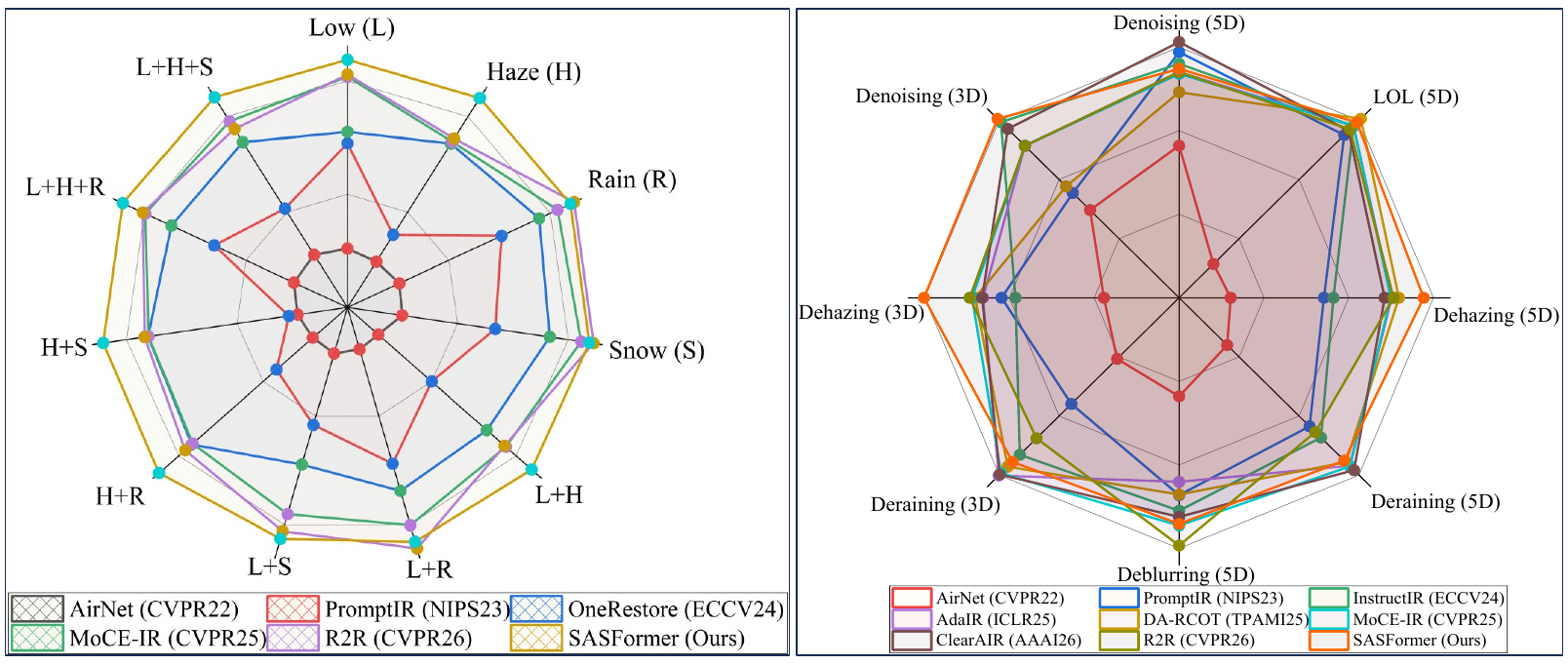} 
	\caption{PSNR comparison with representative methods in the three-degradation and five-degradation settings and on composited degradations.}
	\label{Rader}
\end{figure}

Extensive experiments on 3-, 5-, and 11-degradation settings demonstrate the effectiveness and efficiency of SASFormer, as shown in Fig.~\ref{Rader}. Stage-wise representation analysis further shows that applying adaptive selection to decoding better supports convergence toward a unified representation for clean image reconstruction.
The main contributions of this work are summarized as follows:
\begin{itemize}
     
     \item We reveal distinct representation requirements across encoding and decoding stages in all-in-one image restoration, and accordingly propose a Stage-Aware Selective Attention Framework. It differentiates feature modeling according to stage-specific roles, emphasizing robust representation preservation during encoding and restoration-oriented selective reconstruction during decoding.
    
    \item To enable adaptive selective reconstruction in the decoder, we develop Adaptive Selective Channel Self-Attention (ASCSA) and Adaptive Selective Spatial Self-Attention (ASSSA). ASCSA dynamically identifies informative channel responses through query-channel selection and attention masking, while ASSSA adaptively integrates multi-window spatial attention to capture degradation-dependent spatial contexts.
    
    \item Extensive experiments on 3-, 5-, and 11-degradation settings demonstrate competitive restoration performance and favorable parameter efficiency. Stage-wise representation analysis and ablation studies further support the proposed stage-aware design, while cross-degradation evaluations demonstrate its generalization across diverse degradation scenarios.
    
\end{itemize}

\section{Related Work}
\label{sec:Related-Works}
\paragraph{Task-Specific Image Restoration.}
Image restoration is a fundamental research topic in computer vision. Due to the complexity and inherent uncertainty of degradation processes, traditional methods typically rely on manually designed features and prior knowledge to construct restoration models \cite{li2018joint, he2010single, berman2016non}. With the rapid advancement of deep learning, convolutional neural network (CNN)-based approaches have attracted considerable attention, leading to numerous effective methods for diverse tasks such as image denoising \cite{Denoise_DeamNet,Denoise_pan2023random, Denoise_lin2023unsupervised}, dehazing \cite{Dehaze_chen2020pmhld, Dehaze_qin2020ffa,Dehaze_shao2020domain}, deraining \cite{Derain_chen2021robust, Derain_li2019heavy}, deblurring \cite{Deblur_kupyn2019deblurgan, chen2024unsupervised}, and image super-resolution~\cite{gao2024learning}. However, convolutional neural networks exhibit limitations in modeling long-range dependencies within images. To address this issue, researchers have introduced Transformers into the field of image restoration, including spatial self-attention-based methods such as SwinIR \cite{liang2021swinir}, Uformer \cite{wang2022uformer}, and HAT \cite{chen2023activating}, channel self-attention-based approaches like Restormer \cite{IM_zamir2022restormer}, as well as hybrid attention models such as X-Restormer \cite{chen2024comparative}. These Transformer-based methods are capable of more effectively restoring fine details and clarity in images with complex textures and structures. Complementary efforts include ECFNet~\cite{gao2026emphasizing}, which emphasizes crucial features through spatial and frequency attention, and StarIR~\cite{cui2026starir}, which combines convolutional spatial--frequency representations through multiplicative feature fusion. Although effective for individual image restoration tasks, it remains challenging to simultaneously address diverse requirements within an all-in-one framework.

\paragraph{All-in-One Image Restoration.}
All-in-one methods aim to handle multiple types of image degradation within a unified architecture. Early approaches typically relied on multiple encoders or decoders to distinguish different degradations \cite{chen2021pre, li2020all}, whereas AirNet \cite{all-in-one_AirNet_li2022all} was the first to employ a single encoder–decoder framework, achieving unified image restoration through contrastive learning without requiring degradation priors. PromptIR \cite{potlapalli2023promptir} employs implicit prompts, highlighting the crucial role of prompting mechanisms in all-in-one frameworks. IDR \cite{all-in-one_IDR_zhang2023ingredient} employs a meta-learning-based two-stage approach to model degradation through underlying physical principles. Since then, all-in-one methods have attracted increasing attention, and numerous related approaches \cite{conde2025instructir, 10970105, Perceive-IR-24, zeng2025vision, zamfir2025complexity, yu2024multi, yang2024all, he2026degradation, zhang2025uniuir} have been proposed in recent years. Among them, auxiliary network or training strategy-based methods \cite{all-in-one_AirNet_li2022all, wu2024harmony, conde2025instructir, 10970105, Perceive-IR-24, zeng2025vision, tian2025degradation} have attracted increasing attention and research interest due to their remarkable performance. The core idea of these methods is to employ auxiliary networks or additional training strategies to identify degradation types, and then leverage the obtained degradation priors to guide the restoration of multi-degraded images. 
For example, DFPIR \cite{tian2025degradation} introduces text prompts to provide degradation guidance and incorporates channel perturbation to enhance the fusion of degradation information into feature representations. ClearAIR~\cite{zhang2026clearair} further introduces a segmentation-guided mechanism to explicitly decouple structural regions and degradation components, enhancing restoration consistency and visual quality. AllRestorer~\cite{mao2026all} addresses adverse weather degradations using an all-in-one Transformer block with image--text scene embeddings and adaptive degradation weights.

Among recent all-in-one image restoration studies published in the \emph{IEEE Transactions on Circuits and Systems for Video Technology} (TCSVT), the works most closely related to ours are AIRFormer~\cite{gao2023frequency}, CAPTNet~\cite{gao2024prompt}, MEASNet~\cite{yu2024multi}, PalRestorer~\cite{zheng2026crafting}, and IPT-ILR~\cite{yang2024ipt}. AIRFormer exploits frequency-guided Transformer representations, while CAPTNet employs degradation-specific prompts for decoder guidance. MEASNet performs adaptive expert selection, PalRestorer models degradation-aware task affinities, and IPT-ILR exploits an image-pyramid Transformer with information-loss regularization. These methods introduce adaptivity through frequency representations, prompts, experts, task relationships, or multi-scale architectures. 
In contrast, SASFormer addresses the distinct representation requirements of encoding and decoding through a stage-aware architecture. The encoder employs established hybrid attention for robust representation learning, whereas adaptive channel and spatial selection are specifically developed to meet the restoration-oriented requirements of decoder reconstruction. This distinction separates SASFormer from the above approaches, which primarily improve degradation or task adaptivity without explicitly modeling the different functional requirements of encoding and decoding stages.

Our study focuses on the distinct representation requirements of encoding and decoding under diverse degradations. Accordingly, we propose a stage-aware architecture that differentiates feature modeling across restoration stages, and further develop adaptive channel and spatial selection tailored to decoder reconstruction. This design improves the performance–efficiency trade-off of unified restoration.

\begin{figure*}[tbp]
	\centering
	\includegraphics[width=1\linewidth]{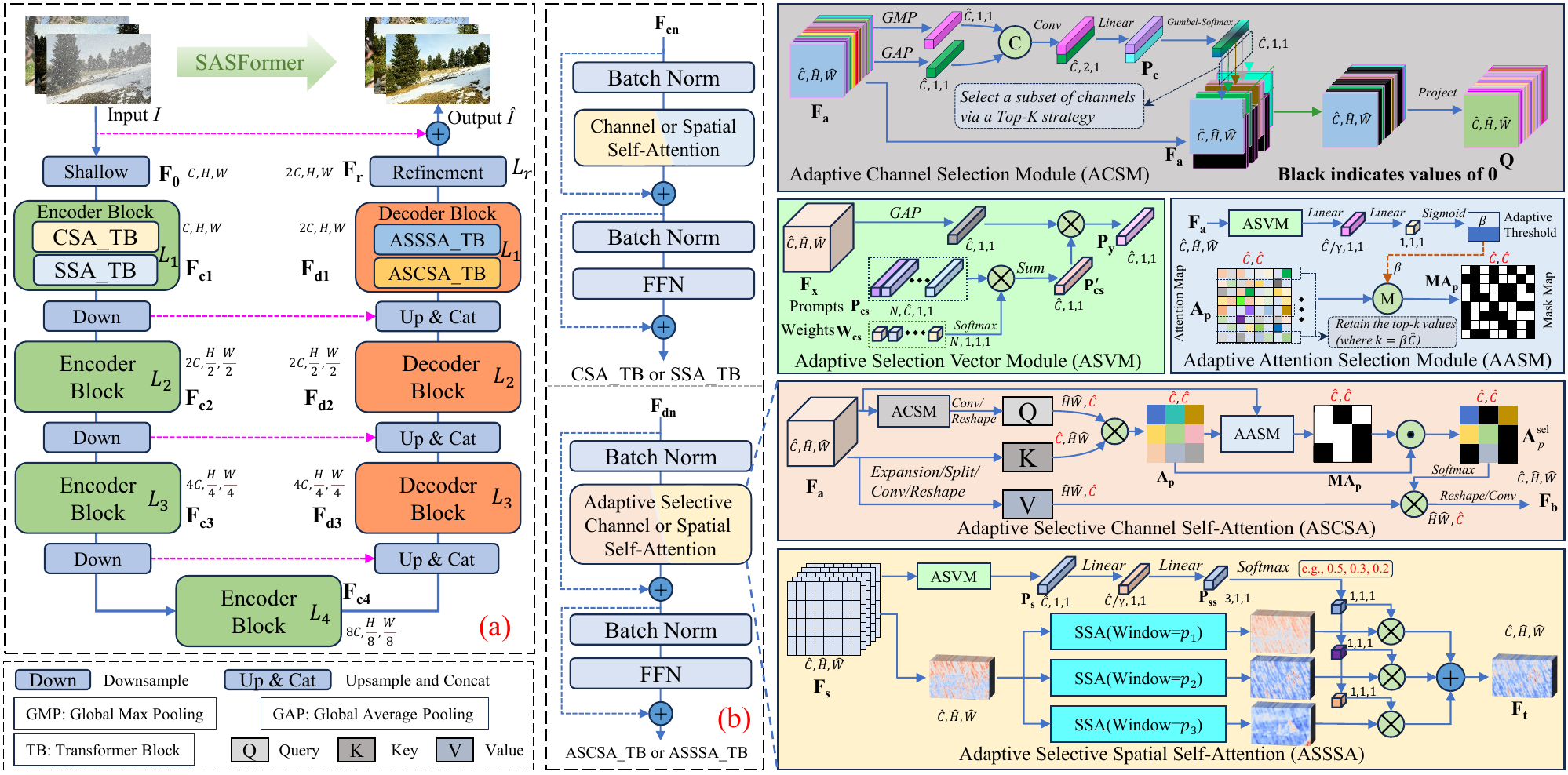} 
	\caption{Overview of the proposed SASFormer architecture. (a) The model adopts a U-shaped backbone with different Transformer blocks in the encoding and decoding stages. (b) The encoder integrates transposed channel self-attention (CSA) \cite{IM_zamir2022restormer} and spatial self-attention (SSA) \cite{chen2023activating}, while the decoder incorporates adaptive selective channel self-attention (ASCSA) and adaptive selective spatial self-attention (ASSSA). ASCSA comprises ACSM and AASM, with ACSM responsible for query-channel selection and AASM performing attention selection using ASVM. ASSSA employs another independent ASVM instance to predict fusion weights for spatial attention outputs at different window sizes.}
	\label{overall-framework}
\end{figure*}

\section{Proposed Method}
\label{Method}

The goal of all-in-one blind image restoration is to learn a unified model $\mathbf{M}$ that restores a clean image $\widehat{I}$ from a degraded input $I$ affected by an unknown degradation $\mathbf{D}$. To coordinate representation learning and selective reconstruction, SASFormer is built upon a stage-aware design principle, which assigns different feature modeling objectives to different network stages. Specifically, the encoder is responsible for learning robust image representations while progressively enhancing degradation-aware feature encoding, whereas the decoder focuses on restoration-oriented feature selection to suppress degradation interference and reconstruct intrinsic image structures.
The overall architecture of SASFormer is illustrated in Fig.~\ref{overall-framework}. The proposed stage-aware design and its key components are detailed in the following subsections.

\subsection{Motivation and Overview Pipeline}
\label{method-motivation}

\begin{figure}[tb]
	\centering
	\includegraphics[width=1\linewidth]{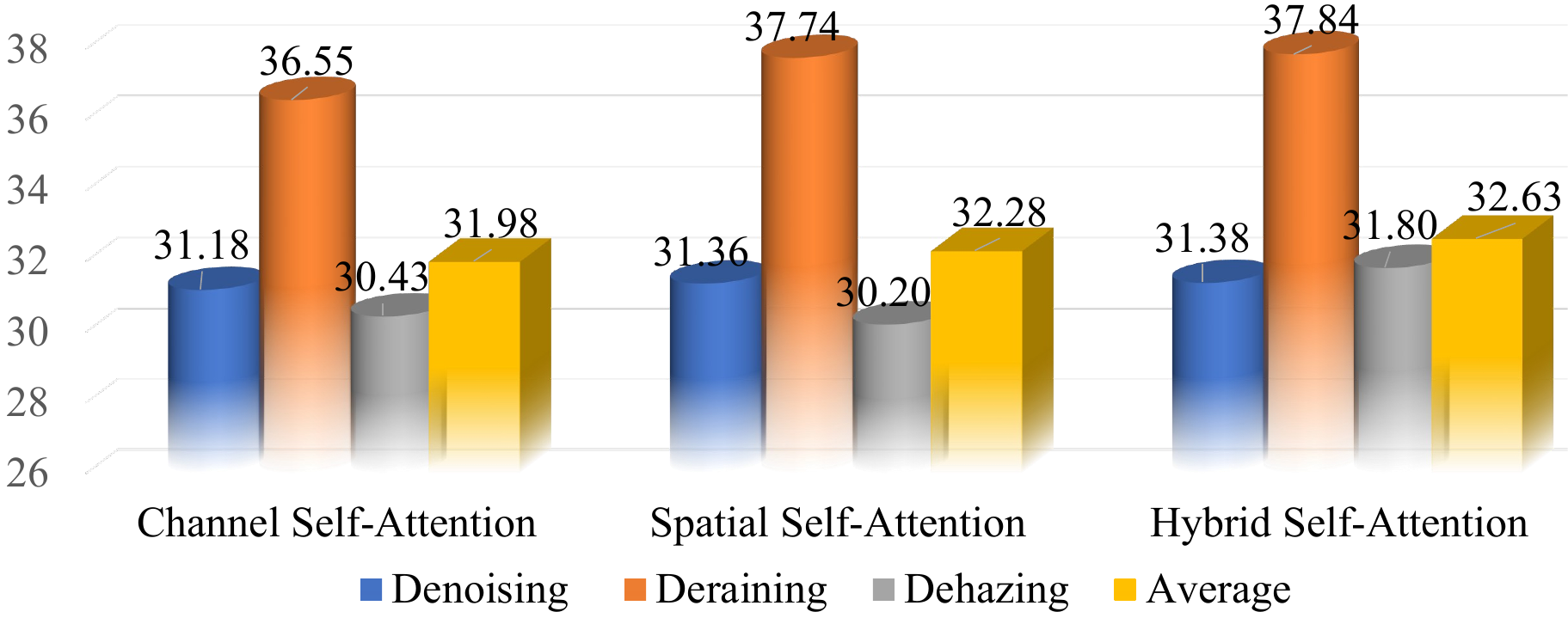}
	\caption{Performance metrics of different self-attention backbones under the all-in-one setting (three degradations) show that hybrid attention achieves the strongest feature extraction and representation capabilities, making it particularly well-suited for all-in-one restoration scenarios.}
	\label{motivation-table}
\end{figure}

\begin{figure}[tb]
	\centering
	\includegraphics[width=1\linewidth]{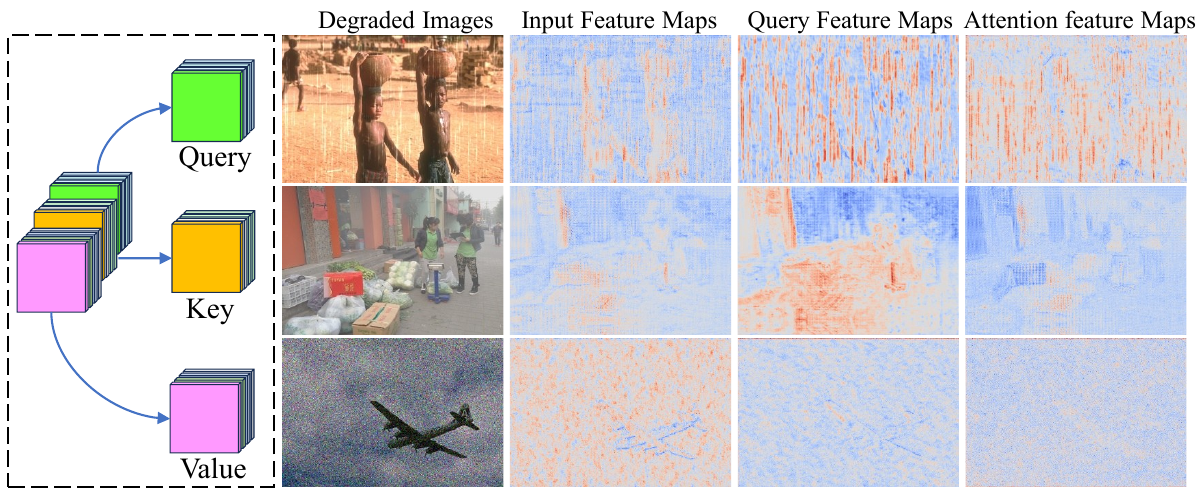} 
	\caption{Feature visualization of the hybrid self-attention method \cite{chen2024comparative}. Under fixed channel grouping, attention responses vary across degradations: haze-related channels are partially suppressed, while rain-related features remain insufficiently attenuated, indicating the limitation of static splitting in all-in-one restoration.}
	\label{motivation-visual}
\end{figure}

\textbf{Motivation.}
Existing all-in-one image restoration methods widely adopt channel, spatial, or hybrid attention mechanisms. As shown in Fig.~\ref{motivation-table}, hybrid attention consistently improves performance over single-type attention across multiple degradations, indicating its effectiveness in capturing complementary representations.
However, as illustrated in Fig.~\ref{motivation-visual}, we observe that such improvements are still limited under a static design paradigm. Specifically, fixed feature partitioning and non-adaptive attention allocation lead to inconsistent restoration behavior across different degradations (e.g., denoising vs. deraining), where a single model cannot simultaneously satisfy diverse representation requirements.

These observations suggest that fixed hybrid attention alone may be insufficient to accommodate degradation-dependent reconstruction requirements, motivating a stage-aware allocation in which established hybrid attention preserves encoder representations and adaptive selective attention is reserved for decoder reconstruction.

\textbf{Overview Pipeline.} Our SASFormer adopts a U-shaped architecture \cite{IM_zamir2022restormer}, as shown in Fig.~\ref{overall-framework}. Given a degraded input ${I} \in {\mathbb{R}^{3 \times H \times W}}$, a $3 \times 3$ convolution first extracts shallow features ${\textbf{F}_\mathbf{0}} \in {\mathbb{R}^{C \times H \times W}}$. These are fed into a four-level encoder–decoder network to produce deep features ${\textbf{F}_\mathbf{r}} \in {\mathbb{R}^{2C \times H \times W}}$. Each encoder and decoder level contains multiple Transformer blocks \cite{chen2024comparative}, with the number of blocks increasing from top to bottom. The encoder gradually reduces spatial resolution and expands channel capacity, yielding a compact latent representation ${\textbf{F}_\mathbf{e4}} \in {\mathbb{R}^{8C \times \frac{H}{8} \times \frac{W}{8}}}$. The decoder reconstructs the high-resolution image with skip connections \cite{ronneberger2015u} fusing encoder and decoder features. Specifically, the encoder integrates hybrid channel and spatial self-attention blocks \cite{IM_zamir2022restormer, chen2023activating} to enhance feature representation, while the decoder employs adaptive selective channel and spatial self-attention to mitigate multi-degradation interference and strengthen intrinsic feature extraction.

\subsection{Encoder: Representation Preservation with Hybrid Attention}
To address the limited global receptive field of CNNs and the quadratic complexity of standard visual Transformers, Restormer~\cite{IM_zamir2022restormer} introduces Channel Self-Attention (CSA), which unfolds features along the channel dimension to compute attention among channels, providing strong global semantic modeling capability. Spatial self-attention \cite{liang2021swinir}, on the other hand, partitions the image into fixed-size spatial windows and computes attention across these windows to capture spatial dependencies, with varying window sizes accommodating the diverse spatial refinement requirements of different degradations. Combining channel and spatial self-attention is particularly suitable for all-in-one scenarios, where different degradations demand varied feature representation capabilities. 
In this work, we adopt CSA and SSA as fundamental building blocks to jointly model complementary global and local representations. These standard attention mechanisms are adopted with minor modifications to their hyperparameter settings for improved parameter efficiency, while preserving their original attention formulations. Within the proposed stage-aware architecture, CSA and SSA are used for robust content-preserving and degradation-aware representation learning in the encoder, while adaptive selective attention is specifically developed to meet the restoration-oriented requirements of decoder reconstruction.

\subsection{Decoder: Restoration-Oriented Selective Attention}
In all-in-one image restoration, the decoder reconstructs high-quality images from encoded features containing both structural information and implicit degradation cues. Since different degradations demand different reconstruction behaviors, uniform feature propagation may be insufficient for restoration-oriented decoding.
To address this limitation, we develop Adaptive Selective Channel Self-Attention (ASCSA) and Adaptive Selective Spatial Self-Attention (ASSSA), which dynamically select restoration-relevant information from the channel and spatial dimensions, respectively. This enables degradation-adaptive reconstruction under diverse degradation conditions.

\subsubsection{Adaptive Selective Channel Self-Attention (ASCSA)}
\label{sec:ASCSA}
Our proposed ASCSA mechanism incorporates two levels of adaptive
selection modules: the \textit{Adaptive Channel Selection Module} (ACSM)
and the \textit{Adaptive Attention Selection Module} (AASM). Specifically,
ACSM adaptively determines whether each channel should participate in
subsequent attention computation, enabling the model to emphasize
intrinsic image representations while suppressing degradation-related
responses. Subsequently, AASM further refines the attention responses by
adaptively selecting informative attention components, thereby reducing
degradation interference.
Given the input feature
$\mathbf{F}_{a}\in\mathbb{R}^{B\times\hat{C}\times\hat{H}\times\hat{W}}$,
ACSM first aggregates global contextual information through global
average pooling (GAP) and global max pooling (GMP). The aggregated
features are then fused and projected by a lightweight network to
generate channel-wise selection logits
$\mathbf{P}_{c}\in\mathbb{R}^{B\times\hat{C}\times2\times1}$,
where the two logits correspond to the ``select'' and ``discard'' states
for each channel, respectively.
Unlike hard Top-$K$-based channel selection strategies that require a
manually predefined selection ratio, ACSM performs adaptive binary
channel selection based on the learned channel-wise logits. Specifically,
the Straight-Through Gumbel-Softmax estimator is employed to generate
channel selection decisions:
\begin{equation}
	\mathbf{S}
	=
	\operatorname{GumbelSoftmax}
	\left(
	\mathbf{P}_{c},
	\tau_{t}^{\mathrm{ch}},
	\operatorname{dim}=2
	\right),
\end{equation}
where $\tau_{t}^{\mathrm{ch}}$ denotes the temperature parameter
controlling the discreteness of the selection process, and
$\operatorname{dim}=2$ indicates that the Gumbel-Softmax operation is
performed over the two selection states for each channel. The resulting
selection tensor
$\mathbf{S}\in\mathbb{R}^{B\times\hat{C}\times2\times1}$
represents the binary channel-selection decisions in the forward pass,
while its underlying continuous Gumbel-Softmax relaxation provides
differentiable gradients in the backward pass.

The Straight-Through estimator adopts hard one-hot channel decisions
during the forward propagation while maintaining differentiable
gradients through the continuous Gumbel-Softmax relaxation during the
backward propagation. In this way, the channel-selection logits can be
jointly optimized with the restoration network without explicitly
differentiating through the discrete channel selection operation.
To progressively sharpen the channel-selection behavior, the
temperature is annealed during training as
\begin{equation}
	\tau_{t}^{\mathrm{ch}}
	=
	\tau_{\mathrm{start}}^{\mathrm{ch}}
	\left(
	\frac{\tau_{\mathrm{end}}^{\mathrm{ch}}}
	{\tau_{\mathrm{start}}^{\mathrm{ch}}}
	\right)^{t/T},
\end{equation}
where $t$ denotes the current training epoch and $T$ represents the
total number of training epochs. The decreasing temperature gradually
reduces the uncertainty of channel selection while retaining hard
decisions in the forward pass.
The final binary channel mask is obtained by extracting the
``select'' state:
\begin{equation}
	\mathbf{M}^{\mathrm{ch}}
	=
	\mathbf{S}_{:,:,0,:}
	\in
	\mathbb{R}^{B\times\hat{C}\times1\times1},
\end{equation}
where the index $0$ denotes the ``select'' state and the index $1$
denotes the ``discard'' state. The selected query features are then
calculated as
\begin{equation}
	\mathbf{F}_{Q}
	=
	\mathbf{F}_{a}
	\odot
	\mathbf{M}^{\mathrm{ch}},
\end{equation}
where $\odot$ denotes element-wise multiplication. The selected
features are subsequently projected through a lightweight convolutional
layer to match the required feature dimension for the following
attention computation.

Importantly, the same Gumbel-Softmax-based channel selection mechanism is employed during both training and inference, ensuring consistent channel selection throughout the entire restoration process. The channel selection is determined by the learned channel-wise selection logits, allowing ACSM to dynamically identify informative channels according to the degradation-aware representations without requiring a manually predefined channel selection ratio.

\begin{figure}[tb]
	\centering
	\includegraphics[width=1\linewidth]{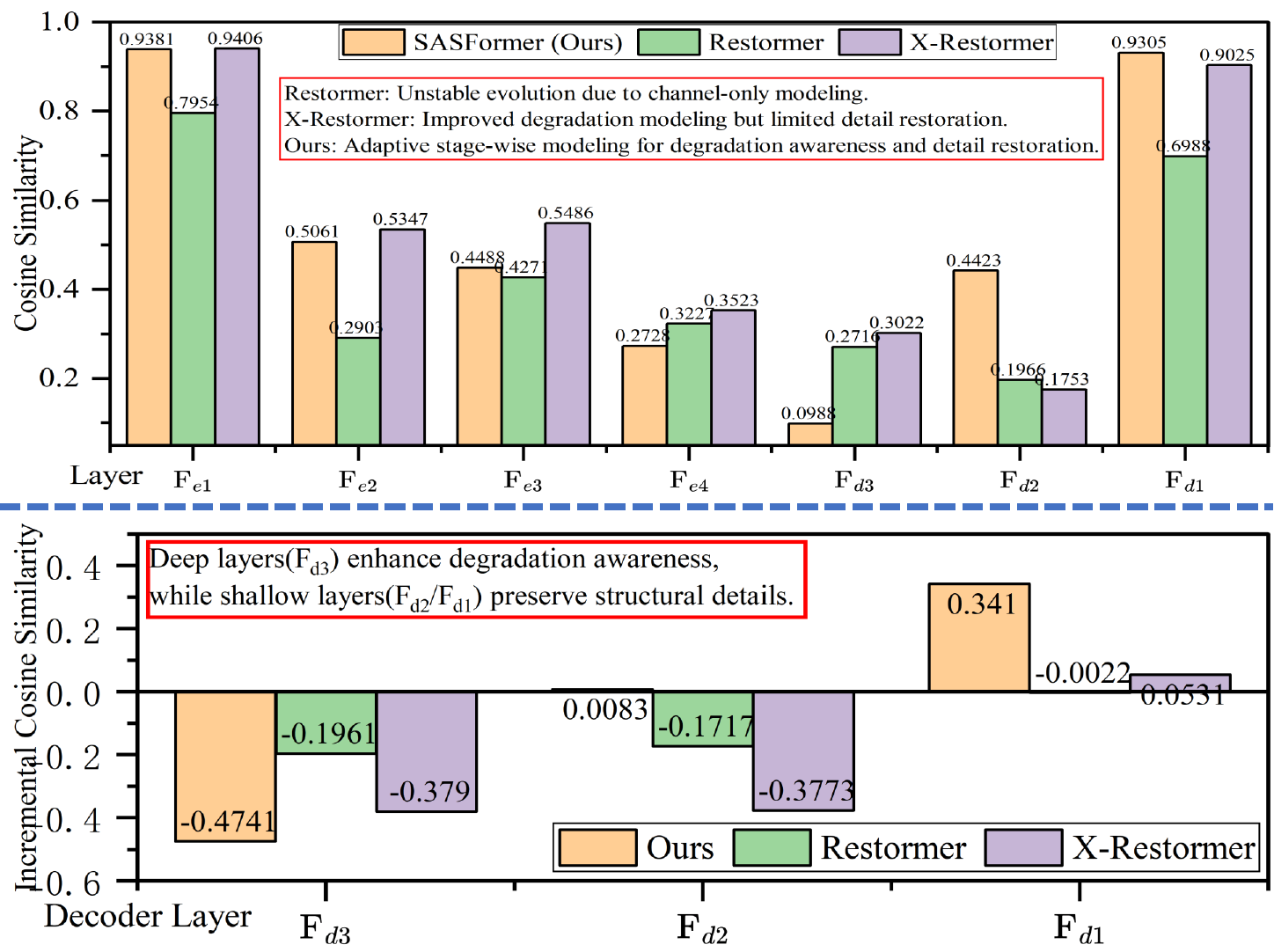} 
	\caption{\textbf{Top}: Comparison of cosine similarity of features extracted from the same clean image under rain and noise degradations on the Rain100L dataset, highlighting representation discrepancies induced by different self-attention mechanisms across stages.
		\textbf{Bottom}: Improvement in cosine similarity after introducing the ASCSA/ASSSA modules, further demonstrating the effectiveness of the proposed selective self-attention in enhancing degraded feature representation in deep layers and strengthening intrinsic structure modeling in shallow layers.}
	\label{cosine-backbone}
\end{figure}

To further mitigate the impact of multiple degradations, motivated by PIP~\cite{li2023prompt}, we propose an AASM. Unlike PIP, which manually sets fixed mask thresholds for all degradation types and levels, we design an \textit{Adaptive Selection Vector Module} (ASVM) that integrates feature information and implicit prompts to automatically determine the mask threshold.
Specifically, the learnable implicit prompts ${{\mathbf{P}}_{\mathbf{cs}}} \in {\mathbb{R}^{N \times \widehat{C} \times 1 \times 1}}$ (where $N$, the number of prompts, is set to 3) are first multiplied by the weights ${{\mathbf{W}}_{\mathbf{cs}}} \in {\mathbb{R}^{N \times 1 \times 1 \times 1}}$, which are independently learnable parameters shared across all images rather than predicted from the input, and then summed to obtain a weighted prompt vector ${\mathbf{P}}_{{\mathbf{cs}}}^\prime \in {\mathbb{R}^{\widehat{C} \times 1 \times 1}}$. Finally, this vector is further modulated by the channel features of ${{\textbf{F}}_{\mathbf{x}}} \in {\mathbb{R}^{\widehat{C} \times \widehat{H} \times \widehat{W}}}$ obtained via GAP to produce ${{\mathbf{P}}_{\mathbf{y}}} \in {\mathbb{R}^{\widehat{C} \times 1 \times 1}}$. The implicit prompts are internally defined learnable tensors shared across all images.
They are initialized with ones and optimized jointly with the network parameters without additional supervision. The entire ASVM process can be described as follows:
\begin{equation}
	{{\mathbf{P}}_\mathbf{y}} = Sum({{\mathbf{P}}_{\mathbf{cs}}} \cdot Softmax({{\mathbf{W}}_{\mathbf{cs}}})) \cdot GAP({{\mathbf{F}}_\mathbf{x}}).
\end{equation}

\begin{figure}[tb]
	\centering
	\includegraphics[width=1\linewidth]{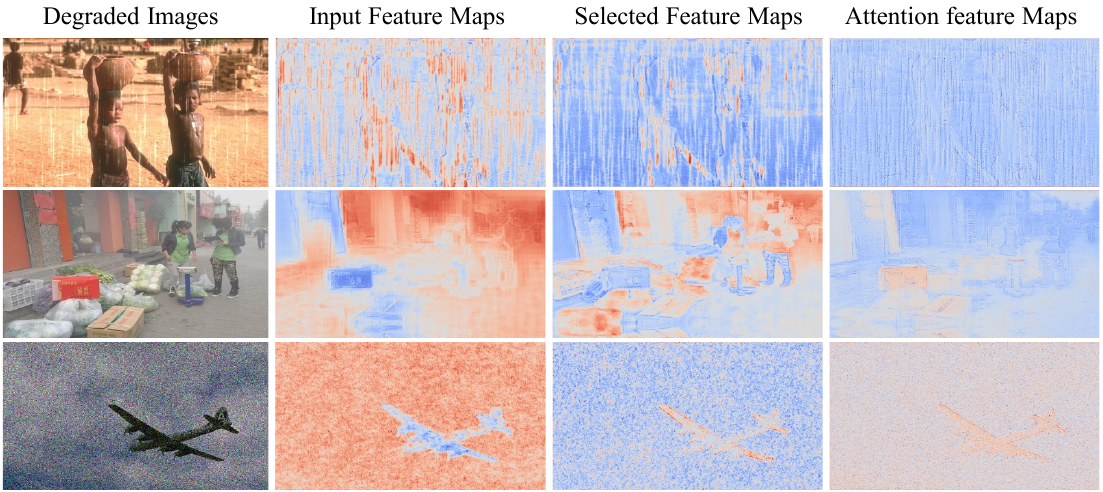}
	\caption{Visualization of features from the adaptive selective channel attention. The selected channels mainly focus on the intrinsic image content, effectively suppressing various degradation features and thereby demonstrating strong robustness across different degradations.}
	\label{fig:Selective-Channel-Visual}
\end{figure}

In AASM, the feature map $\mathbf{F}_a$ and the pre-softmax attention
logits $\mathbf{A}_p\in\mathbb{R}^{\widehat C\times\widehat C}$ are used
to construct a degradation-adaptive attention mask. Specifically,
$\mathbf{F}_a$ is first processed by ASVM and then fed into a learnable
projection network to produce a scalar adaptive selection factor:
\begin{equation}
	\beta
	=
	Sigmoid
	\left(
	f_{\theta}(\operatorname{ASVM}(\mathbf{F}_a))
	\right),
	\qquad
	\beta\in(0,1),
	\label{eq:aasm_beta}
\end{equation}
where $f_{\theta}(\cdot)$ denotes the two-layer projection network.
Rather than directly thresholding the attention values, $\beta$ controls
the number of retained entries in each attention row. We first map it to a
continuous selection budget:
\begin{equation}
	K_s
	=
	1+(\widehat C-1)\beta,
	\qquad
	k
	=
	\lfloor K_s\rfloor,
	\label{eq:aasm_budget}
\end{equation}
where $\lfloor\cdot\rfloor$ denotes the floor operation. Thus, $K_s$ is a
continuous target number of retained entries, whereas $k$ is the integer
number used by the hard Top-$K$ operation. This formulation guarantees
that at least one entry is retained in every attention row.
Let $r_{ij}\in\{0,\ldots,\widehat C-1\}$ denote the descending rank of
$A_{p,ij}$ within the $i$-th row of $\mathbf{A}_p$, where $r_{ij}=0$
corresponds to the largest attention logit. The hard row-wise Top-$K$
mask used in the forward pass is defined as
\begin{equation}
	M^{\mathrm{hard}}_{ij}
	=
	\mathbb{I}\left[r_{ij}<k\right],
	\label{eq:aasm_hard_mask}
\end{equation}
where $\mathbb{I}[\cdot]$ denotes the indicator function, which equals
$1$ when its condition is satisfied and $0$ otherwise. Accordingly,
$M^{\mathrm{hard}}_{ij}=1$ if $A_{p,ij}$ belongs to the Top-$k$ entries
of the $i$-th row, and $M^{\mathrm{hard}}_{ij}=0$ otherwise.
Because the ranking and integer Top-$K$ operations are non-differentiable,
we use a rank-based continuous relaxation during back-propagation:
\begin{equation}
	\widetilde{M}_{ij}
	=
	Sigmoid
	\left(
	\frac{K_s-r_{ij}-0.5}{\tau^{\mathrm{att}}}
	\right),
	\label{eq:aasm_soft_mask}
\end{equation}
where $\tau^{\mathrm{att}}$ is the temperature of the attention-selection
surrogate. The hard and soft masks are coupled through a straight-through
estimator (STE):
\begin{equation}
	\mathbf{M}^{\mathrm{ST}}
	=
	\mathbf{M}^{\mathrm{hard}}
	+
	\widetilde{\mathbf{M}}
	-
	\operatorname{sg}(\widetilde{\mathbf{M}}),
	\label{eq:aasm_ste}
\end{equation}
where $\operatorname{sg}(\cdot)$ denotes the stop-gradient operator.
Therefore, $\mathbf{M}^{\mathrm{ST}}$ equals the exact hard Top-$K$ mask
in the forward pass, while its gradient in the backward pass is given by
the continuous surrogate $\widetilde{\mathbf{M}}$.
The attention-selection temperature is annealed independently from the
channel-selection temperature according to
\begin{equation}
	\tau_t^{\mathrm{att}}
	=
	\tau_{\mathrm{start}}^{\mathrm{att}}
	\left(
	\frac{\tau_{\mathrm{end}}^{\mathrm{att}}}
	{\tau_{\mathrm{start}}^{\mathrm{att}}}
	\right)^{t/T},
	\label{eq:aasm_temperature}
\end{equation}
where $t$ and $T$ denote the current and total training epochs,
respectively. The decreasing temperature
progressively sharpens the relaxed mask around the Top-$K$ boundary,
whereas the forward pass always uses the hard mask.

\begin{figure}[tb]
	\includegraphics[width=1\linewidth]{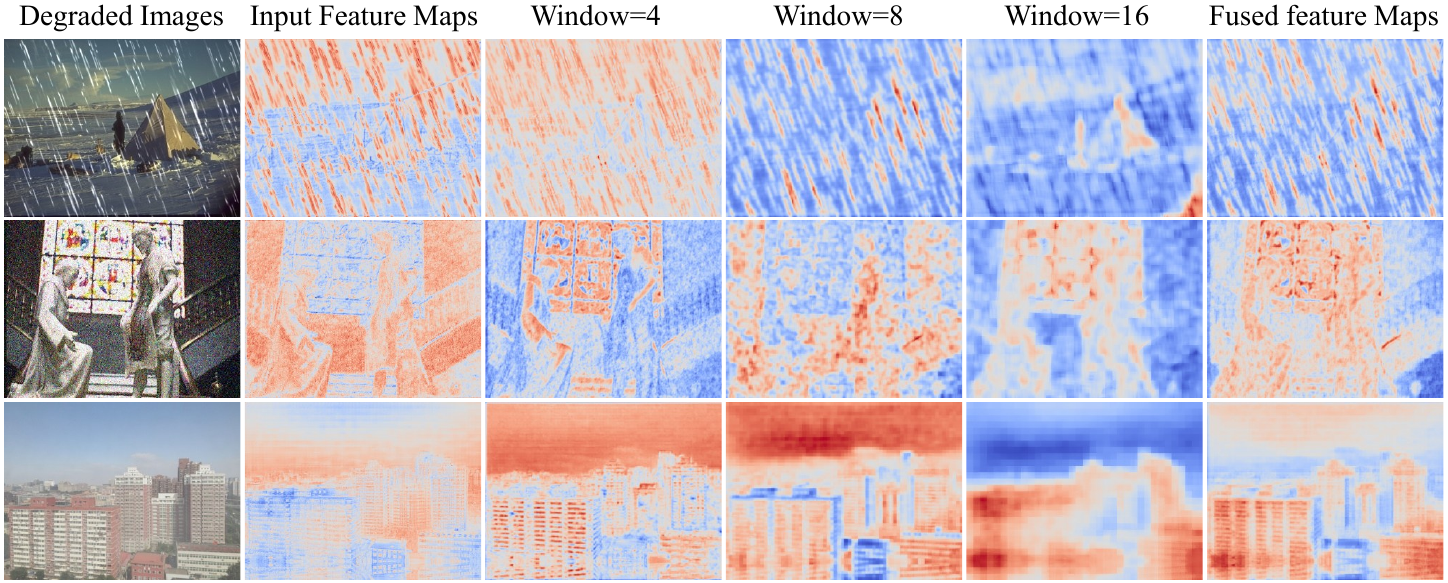}
	\caption{Attention feature maps under different window sizes. Smaller windows primarily capture fine-grained feature details, providing limited suppression of degradation-related features, whereas larger windows focus on broader structural information, yielding stronger suppression of degradation features.}
	\label{fig:Spatial-window-visual}
\end{figure}

Let $\overline{\mathbf{A}}_p=\operatorname{Softmax}(\mathbf{A}_p)$
denote the unmasked attention weights. We obtain the selected attention
weights through differentiable multiplication and row-wise
renormalization:
\begin{equation}
	A^{\mathrm{sel}}_{p,ij}
	=
	\frac{
		\overline{A}_{p,ij}M^{\mathrm{ST}}_{ij}
	}{
		\sum_{\ell=1}^{\widehat C}
		\overline{A}_{p,i\ell}M^{\mathrm{ST}}_{i\ell}
		+\epsilon
	},
	\qquad
	\mathbf{F}_b
	=
	\mathbf{A}^{\mathrm{sel}}_p\widehat{\mathbf{V}},
	\label{eq:aasm_output}
\end{equation}
where $\epsilon$ is a small constant for numerical stability and
$\mathbf{A}^{\mathrm{sel}}_p$ denotes the row-wise normalized attention
weights after adaptive Top-$K$ selection.
For the forward hard mask, Eq.~(\ref{eq:aasm_output}) is algebraically
equivalent to applying Softmax only to the retained Top-$K$ entries.
During back-propagation, the reconstruction loss reaches $\beta$ through
the continuous surrogate $\widetilde{\mathbf{M}}$. With the ranks treated
as fixed in the local STE approximation, the corresponding derivative is
\begin{equation}
	\frac{\partial\widetilde{M}_{ij}}{\partial\beta}
	=
	\frac{\widehat C-1}{\tau^{\mathrm{att}}}
	\widetilde{M}_{ij}
	\left(
	1-\widetilde{M}_{ij}
	\right).
	\label{eq:aasm_gradient}
\end{equation}

\begin{figure}[tb]
	\centering
	\includegraphics[width=1\linewidth]{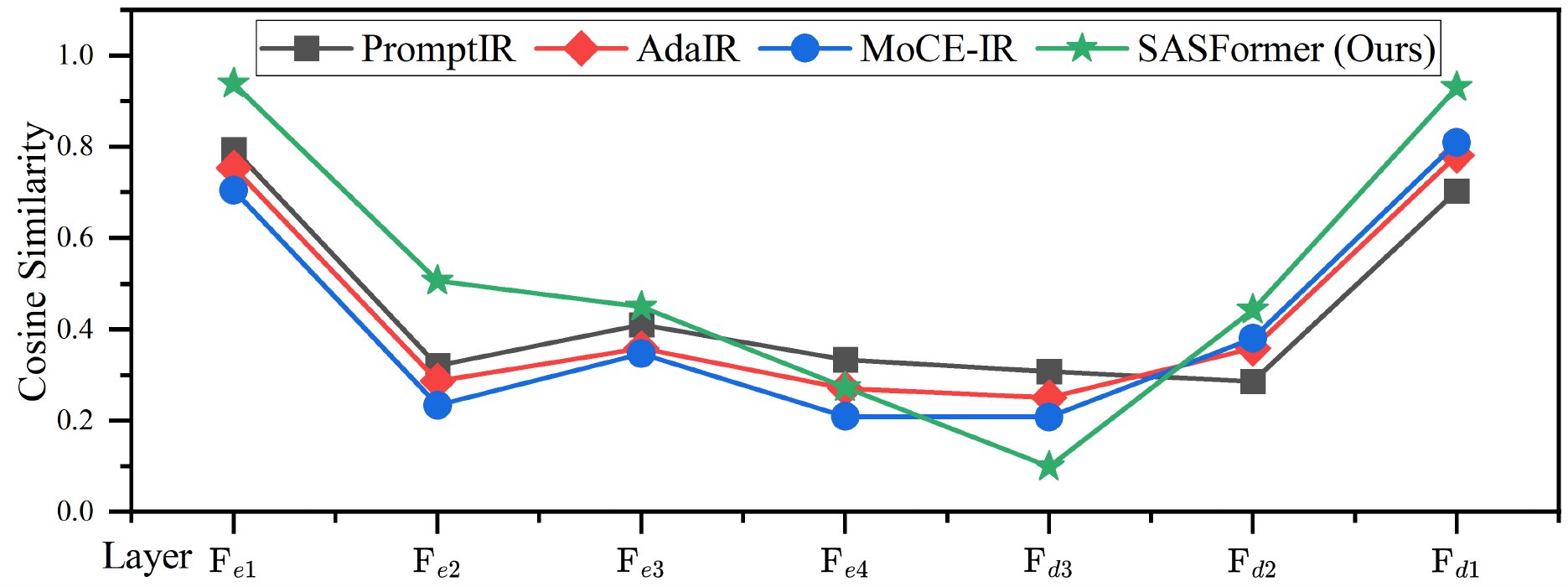} 
	\caption{Stage-wise rain--noise feature similarity of PromptIR, AdaIR,
			MoCE-IR, and SASFormer. Each point is averaged over all 100 matched
			Rain100L test-image pairs. Higher similarity indicates greater feature
			consistency for the same image under rain and Gaussian-noise
			degradations.}
	\label{cosin-comp}
\end{figure}

Hard selection is consistently used in both training and inference. During
training, $\mathbf{M}^{\mathrm{ST}}$ has the same forward value as
$\mathbf{M}^{\mathrm{hard}}$; therefore, the exact row-wise Top-$K$ mask
is applied to the attention computation. The continuous surrogate
$\widetilde{\mathbf{M}}$ is used only to provide the STE gradient during
back-propagation. During inference, we use the same hard Top-$K$ selection
and masked renormalization, without replacing the hard mask by the
continuous surrogate. Consequently, the forward attention-selection
behavior is identical during training and inference.
As shown in Fig.~\ref{fig:Selective-Channel-Visual}, the selected features mainly capture intrinsic image properties, and ASCSA effectively suppresses degradation-related components, demonstrating its ability to mitigate multi-degradation effects.

\subsubsection{Adaptive Selective Spatial Self-Attention (ASSSA)}
\label{sec:ASSSA}
To address the varying spatial attention window size requirements across multiple degradations, we propose an ASSSA mechanism.
For the input feature ${{\textbf{F}}_\mathbf{s}} \in {\mathbb{R}^{\widehat{C} \times \widehat{H} \times \widehat{W}}}$, we first feed it into another ASVM to obtain an adaptive selection vector ${{\mathbf{P}}_{\mathbf{s}}} \in {\mathbb{R}^{\widehat{C} \times 1 \times 1}}$ that integrates both feature and prompt information. This vector is then passed through two Linear layers followed by a Softmax operation to produce the final adaptive selection vector ${{\mathbf{P}}_{\mathbf{ss}}} \in {\mathbb{R}^{3 \times 1 \times 1}}$. The process can be formulated as:
\begin{equation}
	\mathbf{P}_{ss} = Softmax(Linear(ASVM(\mathbf{F}_\mathbf{s}))).
\end{equation}
For $N_w$ SSA modules with different window sizes $w$, their outputs are adaptively weighted and summed using $\mathbf{P}_{ss}$, enabling dynamic feature extraction and representation for images with varying degradation types and levels.  The overall process can be formulated as:
\begin{equation}
	\mathbf{F}_{\mathbf{t}} = \sum_{i=1}^{N_w} \mathbf{P}_{ss}^{(i)} \cdot SSA_i(\mathbf{F}_\mathbf{s}).
\end{equation}
To balance model complexity and representational capacity, we set $N_w = 3$ in this work, with window sizes of $4$, $8$, and $16$, respectively. 
We analyze the adaptive selection vector ${\mathbf{P}}_{\mathbf{ss}}$ in the $F_{d1}$ layer across three degradation datasets, as shown in Tab.~\ref{tab-pss-visual}. A consistent pattern emerges: weak or local degradations favor small windows (\eg, noise15 and rain), while strong or global degradations prefer large windows (\eg, noise50 and haze). As degradation severity increases (from noise15 to noise50), the weight shifts from small to large windows. This confirms that the adaptive selection mechanism dynamically adjusts the receptive field based on degradation strength, underscoring the importance of multi-scale modeling for unified multi-degradation restoration.

\begin{table}[tb]
	\centering
	\caption{Mean weights of the adaptive selection vector ${{\mathbf{P}}_{\mathbf{ss}}}$ in the $F_{d1}$ layer under three degradations. Weak or local degradations (e.g., light noise, local rain) prefer small windows, while strong or global degradations (e.g., heavy noise, haze) favor large windows.}
	\setlength{\tabcolsep}{12.0pt} 
	\label{tab-pss-visual}
	\begin{tabular}{c|ccc}
		\Xhline{1.0pt}
		\multirow{2}{*}{Degradation} & ${{\bf{P}}_{ss}}$ & ${{\bf{P}}_{ss}}$ & ${{\bf{P}}_{ss}}$\\
		& $\left( {w = 4} \right)$ & $\left( {w = 8} \right)$ & $\left( {w = 16} \right)$ \\
		\hline	
		Noise\_15 & \textcolor{red}{0.349}\phantom{x} & 0.314\phantom{x} & 0.337\phantom{x} \\
		Noise\_25 & \textcolor{red}{0.344}\textcolor{cyan}{$\downarrow$} & 0.318\textcolor{green}{$\uparrow$} & 0.338\textcolor{green}{$\uparrow$} \\
		Noise\_50 & 0.337\textcolor{cyan}{$\downarrow$} & 0.325\textcolor{green}{$\uparrow$} & \textcolor{red}{0.339}\textcolor{green}{$\uparrow$} \\
		\hline
		Rain & \textcolor{red}{0.377}\phantom{x} & 0.277\phantom{x} & 0.346\phantom{x} \\
		Haze & 0.338\phantom{x} & 0.317\phantom{x} & \textcolor{red}{0.345}\phantom{x} \\
		\Xhline{1.0pt}
	\end{tabular}
\end{table}

\begin{table*}[htbp]
	\caption{Comparison to state-of-the-art on three degradations (3D). PSNR (dB, ↑) and SSIM (↑) metrics are reported on the full RGB images. Our method achieves competitive task-specific results, the best average PSNR, and a favorable performance-parameter trade-off. (*) denotes methods that leverage foundation models (e.g., CLIP, ResNet, or DINO) or custom-designed auxiliary networks. The \textcolor{red}{best} and \textcolor{blue}{second-best} performances are highlighted.}
	\label{all-in-one-3T}
	\centering
	\setlength{\tabcolsep}{6.5pt} 
	\renewcommand{\arraystretch}{1.0}  
	\begin{tabular}{c|c|ccccc|c|c} 
		\Xhline{1.0pt}  
		\multirow{2}{*}{Types} &\multirow{2}{*}{Method}  & Dehazing  & Deraining & \multicolumn{3}{c|}{Denoising on CBSD68 dataset}  &  \multirow{2}{*}{Average} &  Params.    \\
		\cline{5-7} 
		& & on SOTS  & on Rain100L & $\sigma = 15$& $\sigma = 25$ & $\sigma = 50$ & \phantom{}  &  (M)          \\
		\hline 
		
		\multirow{4}{*}{\rotatebox{90}{General}}&Restormer \cite{IM_zamir2022restormer} (CVPR'22)  &30.43/0.975&36.55/0.975&33.84/0.931&31.18/0.885&27.90/0.790&31.98/0.911 & 26.13\\ 
		
		&NAFNet \cite{IM_NAFNetchen2022simple} (ECCV'22)  &24.11/0.928&33.64/0.956&33.03/0.918&30.47/0.865&27.12/0.754&29.67/0.844 & 17.11\\
		
		&FSNet \cite{cui2023image} (TPAMI'23)  &29.14/0.968&35.61/0.969&33.81/0.930&30.84/0.872&27.69/0.792&29.67/0.844 & 13.28\\
		
		&MambaIR \cite{guo2025mambair} (ECCV'24)  &29.57/0.970&35.42/0.969&33.88/0.931&30.95/0.874&27.74/0.793&31.51/0.907 & 26.78\\
		\hline %
		 
		\multirow{13}{*}{\rotatebox{90}{All-in-One}}&AirNet \cite{all-in-one_AirNet_li2022all} (CVPR'22)  &27.94/0.962&34.90/0.967&33.92/0.933&31.26/0.888&28.00/0.797&31.20/0.910 & 8.93\\

		&IDR \cite{all-in-one_IDR_zhang2023ingredient} (CVPR'23)  &29.87/0.970&36.03/0.971&33.89/0.931&31.32/0.884&28.04/0.798&31.83/0.911 & 15.34\\ 
		
		&PromptIR \cite{potlapalli2023promptir} (NeurIPS'23)  &30.58/0.974&36.37/0.972&33.98/0.933&31.31/0.888&28.06/0.799&32.06/0.913 & 35.59\\ 
		
		&InstructIR* \cite{conde2025instructir} (ECCV'24)  &30.22/0.959&37.98/0.978& 34.15/0.933& 31.52/0.890& 28.30/0.804&32.43/0.913 &15.80 \\
		
		&DPPD (PromptIR) \cite{wu2025learning} (TIP'25) &31.10/\textcolor{blue}{0.981}&37.74/0.982& 34.12/\textcolor{red}{0.941}& \textcolor{red}{31.80}/0.891& 28.17/\textcolor{red}{0.809}&32.59/\textcolor{red}{0.921} &35.59 \\ 
		
		&DA-RCOT* \cite{10970105} (TPAMI'25) &31.26/0.977&38.36/0.983& 33.98/0.934& 31.33/0.890& 28.10/0.801&32.60/0.917 &40.90 \\
		
		&Perceive-IR* \cite{Perceive-IR-24} (TIP'25)  &30.87/0.975&38.29/0.980& 34.13/0.934& \textcolor{blue}{31.53}/0.890& \textcolor{red}{28.31}/0.804&32.63/0.917 &42.02 \\
		
		&AdaIR \cite{cui2025adair} (ICLR'25) &31.06/0.980&38.64/0.983& 34.12/0.935& 31.45/0.892& 28.19/0.802&32.69/0.918 &28.77 \\
		
		&VLU-Net* \cite{zeng2025vision} (CVPR'25) &30.71/0.980&\textcolor{blue}{38.93}/\textcolor{blue}{0.984}& 34.13/0.935& 31.48/0.892& 28.23/0.804&32.70/0.919 &35.35 \\
		
		&MEASNet* \cite{yu2024multi} (TCSVT'25) &\textcolor{blue}{31.61}/\textcolor{blue}{0.981}&\textcolor{red}{39.00}/\textcolor{red}{0.985}& 34.12/0.935& 31.46/0.892& 28.19/0.803&\textcolor{blue}{32.85}/0.919 &31.20 \\
		
		&MoCE-IR \cite{zamfir2025complexity} (CVPR'25) &31.34/0.979&38.57/\textcolor{blue}{0.984}& 34.11/0.932& 31.45/0.888& 28.28/0.800&32.73/0.917 &25.35 \\
		
		&ClearAIR* \cite{zhang2026clearair} (AAAI'26) &31.08/\textcolor{blue}{0.981}&38.61/\textcolor{blue}{0.984}& \textcolor{red}{34.18}/0.935& 31.50/0.891& \textcolor{red}{28.31}/0.804&32.74/0.919 &31 \\

		&R2R \cite{wang2026retrieve} (CVPR'26)
		&31.40/0.977&37.46/0.980&34.10/\textcolor{blue}{0.936}&31.45/\textcolor{red}{0.895}&28.22/0.806&32.53/0.918&19.70\\
		
		\hline
		
		&SASFormer (Ours) 
		&\textcolor{red}{32.58}/\textcolor{red}{0.982}&38.21/0.982&\textcolor{blue}{34.17}/\textcolor{blue}{0.936}& \textcolor{blue}{31.53}/\textcolor{blue}{0.894}        &\textcolor{blue}{28.28}/\textcolor{blue}{0.808}&\textcolor{red}{32.95}/\textcolor{blue}{0.920} &23.69\\	
		\Xhline{1.0pt}  %
	\end{tabular}
\end{table*}

\begin{table*}[tb]
	\caption{Comparison to state-of-the-art on five degradations (5D). PSNR (dB, ↑) and SSIM (↑) metrics are reported on the full RGB images. (*) denotes methods that leverage foundation models (e.g., CLIP, ResNet, or DINO) or custom-designed auxiliary networks. The \textcolor{red}{best} and \textcolor{blue}{second-best} performances are highlighted.} 
	\label{all-in-one-5T}
	\centering
	\setlength{\tabcolsep}{5.5pt} 
	\renewcommand{\arraystretch}{1.0}  
	\begin{tabular}{c|c|ccccc|c|c} 
		\Xhline{1.0pt}  
		
		\multirow{2}{*}{Types}&\multirow{2}{*}{Method} & Dehazing & Deraining & Denoising& Deblurring & Low-light Enh.& \multirow{2}{*}{Average} & Params.\\
		&\phantom{}& on SOTS  & on Rain100L  & on CBSD68  & on GoPro  & on LOL  & \phantom{} &(M)\\
		\hline 
		
		\multirow{5}{*}{\rotatebox{90}{General}}&MIRNetV2 \cite{zamir2022learning} (TPAMI'22)  &24.03/0.927&33.89/0.954&30.97/0.881&26.30/0.799&21.52/0.815&27.34/0.875&5.86 \\

		&Restormer \cite{IM_zamir2022restormer} (CVPR'22) &24.09/0.927&34.81/0.962&31.49/0.884&27.22/0.829&20.41/0.806&27.60/0.881 &26.13 \\ 
		
		&NAFNet \cite{IM_NAFNetchen2022simple} (ECCV'22)   &25.23/0.939&35.56/0.967&31.02/0.883&26.53/0.808&20.49/0.809&27.76/0.881 &17.11 \\ 
		
		&FSNet \cite{cui2023image} (TPAMI'23)  &25.53/0.943&36.07/0.968&31.33/0.883&28.32/0.869&22.29/0.829&28.71/0.898 & 13.28\\
		
		&MambaIR \cite{guo2025mambair} (ECCV'24)  &25.81/0.944&36.55/0.971&31.41/0.884&28.61/0.875&22.49/0.832&28.97/0.901 & 26.78\\
		\hline 
		
		\multirow{14}{*}{\rotatebox{90}{All-in-One}}&Transweather \cite{valanarasu2022transweather} (CVPR'22) &21.32/0.885&29.43/0.905&29.00/0.841&25.12/0.757&21.21/0.792&25.22/0.836 & 37.93 \\ 
		
		&AirNet \cite{all-in-one_AirNet_li2022all} (CVPR'22)   &21.04/0.884&32.98/0.951&30.91/0.882&24.35/0.781&18.18/0.735&25.49/0.846 & 8.93 \\ 
		
		& IDR \cite{all-in-one_IDR_zhang2023ingredient} (CVPR'23)      &25.24/0.943&35.63/0.965&\textcolor{red}{31.60}/0.887&27.87/0.846&21.34/0.826&28.34/0.893 & 15.34 \\
		
		&PromptIR \cite{potlapalli2023promptir} (NeurIPS'23)  &26.54/0.949&36.37/0.970&31.47/0.886&28.71/0.881&22.68/0.832&29.15/0.904 & 35.59\\   
		
		&InstructIR* \cite{conde2025instructir} (ECCV'24)      &27.10/0.956&36.84/0.973 &31.40/0.887&29.40/0.886&23.00/0.836&29.55/0.907 & 15.80 \\
		
		&Perceive-IR* \cite{Perceive-IR-24} (TIP'25)  &28.19/0.964&37.25/0.977&31.44/0.887&29.46/0.886&22.88/0.833&29.84/0.909 & 42.02\\
		
		&DPPD (PromptIR) \cite{wu2025learning} (TIP'25)  &30.31/\textcolor{blue}{0.980}&37.32/0.980&31.33/0.885&28.74/0.875&22.73/0.846&30.09/0.913 & 35.59\\
		
		&VLU-Net* \cite{zeng2025vision} (CVPR'25) &30.84/\textcolor{blue}{0.980}&\textcolor{red}{38.54/0.982}& 31.43/\textcolor{red}{0.891}& 27.46/0.840& 22.29/0.833&30.11/0.905 &35.35 \\
		
		&AdaIR \cite{cui2025adair} (ICLR'25)  &30.53/0.978&38.02/\textcolor{blue}{0.981}&31.35/0.889&28.12/0.858&23.00/0.845&30.20/0.910 & 28.77\\
		
		&DA-RCOT \cite{10970105} (TPAMI'25) &30.96/0.975&37.87/0.980& 31.23/0.888& 28.68/0.872& \textcolor{red}{23.25}/0.836&30.40/0.911 &40.90 \\
		
		&MEASNet* \cite{yu2024multi} (TCSVT'25) &\textcolor{blue}{31.05}/\textcolor{blue}{0.980}&\textcolor{red}{38.32}/\textcolor{red}{0.982}& 31.40/0.888& 29.41/0.890& 23.00/0.845&\textcolor{blue}{30.64}/0.917 &31.20 \\
		
		&MoCE-IR \cite{zamfir2025complexity} (CVPR'25) &30.48/0.974&\textcolor{blue}{38.04}/\textcolor{red}{0.982}&31.34/0.887&\textcolor{blue}{30.05}/\textcolor{blue}{0.899}&23.00/\textcolor{red}{0.852}&30.58/\textcolor{blue}{0.919} & 25.35\\
		
		&ClearAIR* \cite{zhang2026clearair} (AAAI'26) &30.12/0.978&38.20/\textcolor{red}{0.982}& \textcolor{blue}{31.53}/0.888& 29.67/0.887& 22.83/0.846&30.45/0.916 &31 \\
		
		&R2R \cite{wang2026retrieve}(CVPR'26)
		&30.64/0.974&36.61/0.975&31.35/\textcolor{red}{0.891}&\textcolor{red}{30.93}/\textcolor{red}{0.911}&22.88/0.856&30.48/\textcolor{red}{0.921}&19.70\\
		
		\hline							
		&SASFormer (Ours) 
		&\textcolor{red}{32.42}/\textcolor{red}{0.981}&37.80/0.980&31.37/\textcolor{blue}{0.890}&29.98/0.895&\textcolor{blue}{23.12}/\textcolor{blue}{0.851}&\textcolor{red}{30.94}/\textcolor{blue}{0.919} & 23.69 \\	
		
		\Xhline{1.0pt} 
	\end{tabular}				
\end{table*}

\subsection{Analysis of Stage-wise Representation Evolution}
\label{sec:stage_representation}
We analyze how representations of the same image evolve under different
degradations. Using all 100 paired samples in the Rain100L test set, we
use the paired rainy observation and generate a noisy observation from
the same ground-truth image by adding Gaussian noise with
$\sigma=50$. Thus, each test sample provides a matched rain--noise
pair:
\begin{equation}
	I_i^{\mathrm{gt}} \rightarrow
	\left\{
	I_i^{\mathrm{rain}},
	I_i^{\mathrm{noise}}
	\right\}.
\end{equation}
The ground-truth image is used only to construct the noisy observation;
each restoration model receives only the corresponding degraded image.

For the $l$-th stage, we compute the cosine similarity between feature
maps extracted from the matched rain--noise pair:
\begin{equation}
	s_l(i) =
	\operatorname{cos}\left(
	\operatorname{vec}(F_l(I_i^{\mathrm{rain}})),
	\operatorname{vec}(F_l(I_i^{\mathrm{noise}}))
	\right),
\end{equation}
where $F_l(\cdot)$ denotes the feature map at stage $l$, and
$\operatorname{vec}(\cdot)$ directly flattens the complete
$C\times H\times W$ feature map without global average pooling. The
stage-wise similarity is averaged over all test images:
\begin{equation}
	S_l^{\mathrm{rain,noise}} =
	\frac{1}{N}\sum_{i=1}^{N}s_l(i).
	\label{eq:rain_noise_similarity}
\end{equation}
Within this matched-content setting, a decrease in similarity is
consistent with more degradation-dependent feature responses, whereas
an increase in later decoder stages is consistent with the re-alignment
of representations for restoration. We use this measure to characterize
representation dynamics rather than as a direct restoration metric.

Fig.~\ref{cosine-backbone} compares the stage-wise trajectories of
Restormer~\cite{IM_zamir2022restormer}, X-Restormer~\cite{chen2024comparative},
and SASFormer. Restormer and X-Restormer exhibit non-monotonic
representation changes across the encoder--decoder pipeline. In
contrast, SASFormer presents a clearer encoder--decoder transition:
the similarity decreases overall in the encoder and increases again in
the later decoder stages. This behavior is consistent with the proposed
stage-aware design principle. Namely, the encoder retains
degradation-dependent cues instead of prematurely aligning distinct
degradations, whereas the later decoder stages progressively form more
consistent restoration-oriented representations.
The bottom part of Fig.~\ref{cosine-backbone} further illustrates the
effect of introducing ASCSA and ASSSA. The selective modules make the
encoder--decoder transition more evident, suggesting that adaptive
channel and spatial selection helps preserve complementary degradation
and content cues before reconstruction. Rather than employing a uniform
feature modeling strategy throughout the network, SASFormer adaptively
exploits these cues during decoding to suppress degradation interference
and recover intrinsic image structures.
Figs.~\ref{fig:Selective-Channel-Visual} and
\ref{fig:Spatial-window-visual} provide qualitative insight into the
two selective modules. ASCSA assigns higher responses to channels
associated with intrinsic image structures while attenuating less
relevant degradation-related responses. ASSSA combines local and
broader contextual information through adaptive spatial-window
selection. Smaller windows preserve local details, whereas larger
windows capture broader structural context and suppress degradation
artifacts.

\begin{table*}[tb]
	\caption{Comparison with the latest all-in-one state-of-the-art under composited degradations. PSNR (dB, ↑) and SSIM (↑) metrics are reported on the full RGB images. MoCE-IR \cite{zamfir2025complexity} provides training results only for the lightweight version (MoCE-IR-S).``—'' indicates that the metric is not reported.}
	\label{cdd-11-tab_all-in-one}
	\centering
	\setlength{\tabcolsep}{2.5pt} 
	\renewcommand{\arraystretch}{1.1} 
	\fontsize{7.5}{9}\selectfont
	\begin{tabular}{c|ccccccccccc|c} 
		\Xhline{1.0pt} 
		Method& Low (L) & Haze (H) & Rain (R) &  Snow (S)&  L+H & L+R & L+S & H+R & H+S & L+H+R & L+H+S& Avg.\\
		\hline
		WeatherDiff \cite{ozdenizci2023restoring}&23.58/.763&21.99/.904&24.85/.885&24.80/.888&21.83/.756&22.69/.730&22.12/.707&21.25/.868&21.99/.868&21.23/.716&21.04/.698&22.49/.799\\
		AirNet \cite{all-in-one_AirNet_li2022all}& 24.83/.778&24.21/.951&26.55/.891&26.79/.919&23.23/.779&22.82/.710&23.29/.723&22.21/.868&23.29/.901&21.80/.708&22.24/.725&23.75/.814\\
		PromptIR \cite{potlapalli2023promptir}&26.32/.805&26.10/.969&31.56/.946&31.53/.960&24.49/.789&25.05/.771&24.51/.761&24.54/.924&23.70/.925&23.74/.752&23.33/.747&25.90/.850\\
		WGWSNet \cite{zhu2023learning}&24.39/.774&27.90/.982&33.15/.964&34.43/.973&24.27/.800&25.06/.772&24.60/.765&27.23/.955&27.65/.960&23.90/.772&23.97/.771&26.96/.863\\
		OneRestore \cite{guo2024onerestore}&26.48/.826&32.52/\textcolor{blue}{.990}&33.40/.964&34.31/.973&25.79/.822&25.58/.799&25.19/.789&29.99/.957&30.21/.964&24.78/.788&24.90/.791&28.47/.878\\	
		
		DPPD \cite{wu2025learning}& 27.22/---&31.08/---&34.53/---&35.48/---&\textcolor{blue}{26.29}/---&26.36/---&26.11/---&29.87/---&29.64/---&25.30/---&25.37/---&28.84/---\\
		
		MoCE-IR-S \cite{zamfir2025complexity}& 27.26/.824&32.66/\textcolor{blue}{.990}&34.31/\textcolor{blue}{.970}&35.91/\textcolor{blue}{.980}&26.24/.817&26.25/.800&26.04/.793&29.93/\textcolor{blue}{.964}&30.19/\textcolor{blue}{.970}&25.41/.789&\textcolor{blue}{25.39}/.790&29.05/.881\\
		
		R2R \cite{wang2026retrieve}& \textcolor{blue}{27.29}/\textcolor{red}{.836} &32.88/\textcolor{red}{.992}&\textcolor{red}{35.12}/\textcolor{red}{.971} &\textcolor{red}{36.54}/\textcolor{red}{.981}&26.21/\textcolor{red}{.833}&\textcolor{red}{26.70}/\textcolor{red}{.816}&\textcolor{blue}{26.33}/\textcolor{red}{.809}&\textcolor{blue}{30.41}/\textcolor{red}{.966}&\textcolor{blue}{30.37}/\textcolor{red}{.971}&\textcolor{blue}{25.47}/\textcolor{red}{.804}&25.21/\textcolor{red}{.803} & \textcolor{blue}{29.32}/\textcolor{red}{.889}\\
		
		\hline
		
		SASFormer (Ours)&\textcolor{red}{27.50}/\textcolor{blue}{.832}&\textcolor{red}{35.72}/\textcolor{red}{.992}&\textcolor{blue}{34.94}/.969&\textcolor{blue}{36.33}/.978&\textcolor{red}{26.85}/\textcolor{blue}{.827}&\textcolor{blue}{26.57/.808}&\textcolor{red}{26.46}/\textcolor{blue}{.801}&\textcolor{red}{32.08}/\textcolor{red}{.966}&\textcolor{red}{32.29}/\textcolor{red}{.971}&\textcolor{red}{25.96}/\textcolor{blue}{.799}&\textcolor{red}{25.96}/\textcolor{blue}{.797}&\textcolor{red}{30.06}/\textcolor{blue}{.885}\\

		\Xhline{1.0pt} 
	\end{tabular}			
\end{table*}

\begin{table}[tbp]
	\centering
	
	\caption{Cross-degradation generalization results on additional test datasets. PSNR (dB, ↑) and SSIM (↑) are reported on full RGB images. The model is trained under five degradation types and directly evaluated on degradation-specific datasets. Compared with MoCE-IR, our method improves the average PSNR by 1.60 dB, highlighting its superior cross-degradation generalization capability.}
	\setlength{\tabcolsep}{4.0pt} 
	\label{tab-generalization}
	\begin{tabular}{c|cccc} 
		\Xhline{1.0pt} 
		\multirow{2}{*}{Method}&  Deraining &  Deblurring & Low-light& \multirow{2}{*}{Average} \\
		\phantom{}&  Rain100H  &  HIDE  & LOLv2-Real  & \phantom{} \\
		
		\hline 
		IDR \cite{all-in-one_IDR_zhang2023ingredient}  & 11.32/0.397 &16.83/0.621 & 17.61/0.697 &15.25/0.572 \\
		
		Restormer \cite{IM_zamir2022restormer}  & 14.50/0.464 &24.42/0.781 & 26.12/0.877 & 21.68/0.707 \\
		
		PromptIR \cite{potlapalli2023promptir}  & 14.28/0.444 &24.49/0.762 & 26.70/0.870 & 21.82/0.692 \\
		
		AdaIR \cite{cui2025adair}  & 14.13/0.438 &24.43/0.771 & 26.54/0.867 & 21.70/0.692 \\
		MoCE-IR \cite{zamfir2025complexity}  & 14.97/0.472 & 28.51/0.879 & 27.27/0.907 & 23.58/0.753 \\
		\hline
		SASFormer  & \textbf{15.74/0.499} & \textbf{28.06/0.871} & \textbf{31.74/0.915} & \textbf{25.18/0.762} \\
		\Xhline{1.0pt}			
	\end{tabular}
\end{table}

\begin{table}[tbp]
	\centering
	\caption{Zero-shot generalization evaluation. We evaluate the zero-shot generalization performance using model weights trained on three degradation types, and directly test them on the DPDD \cite{ai2024multimodal}, GoPro \cite{nah2017deep}, and HIDE \cite{shen2019human} datasets. All results are reported on RGB images using PSNR (dB, ↑), SSIM (↑), and LPIPS (↓) as quantitative evaluation metrics.}
	\setlength{\tabcolsep}{1.0pt} 
	\label{tab-zeroshot}
	\begin{tabular}{c|ccc} 
		\Xhline{1.0pt}  
		\multirow{2}{*}{Method} &  DPDD & GoPro& HIDE\\
		\phantom{}&     PSNR/SSIM/LPIPS  & PSNR/SSIM/LPIPS  & PSNR/SSIM/LPIPS\\
		
		\hline 
		AirNet \cite{all-in-one_AirNet_li2022all}  & 20.17/0.674/0.376  & 21.95/0.748/0.330 &20.68/0.731/0.350 \\
		
		PromptIR \cite{potlapalli2023promptir}  & 21.76/0.673/0.386  & 22.15/0.749/0.332 & 22.78/0.740/0.333 \\
		
		MoCE-IR \cite{zamfir2025complexity}   & 21.83/0.678/0.372 & 22.50/0.758/0.326 & 22.92/0.743/0.334 \\
		
		\hline
		SASFormer   &\textbf{22.43/0.704/0.372} & \textbf{23.17/0.755/0.327} & \textbf{23.07/0.740/0.331} \\
		\Xhline{1.0pt}			
	\end{tabular}
\end{table}

The preceding comparison analyzes SASFormer against general restoration
backbones. To directly compare representative all-in-one restoration
methods under the same matched rain--noise inputs, we further evaluate
PromptIR, AdaIR, MoCE-IR, and SASFormer. For each method, feature maps
are extracted at semantically corresponding encoder, bottleneck, and
decoder stages, matched by their relative encoder--decoder position and
spatial resolution. All methods are evaluated on the same fixed 100
rain--noise pairs using the metric in Eq.~(\ref{eq:rain_noise_similarity}).
Fig.~\ref{cosin-comp} presents the resulting stage-wise trajectories.
SASFormer exhibits a coherent divergence-to-late-convergence pattern:
its similarity decreases consistently throughout the encoder, indicating
that degradation-dependent cues are retained rather than prematurely
aligned. In contrast, PromptIR, AdaIR, and MoCE-IR all show a local
re-increase in intermediate encoder stages, resulting in less regular
encoder-stage representation evolution.
After the bottleneck, SASFormer further differentiates the matched
rain--noise features in the early decoder and progressively re-aligns
them in the subsequent decoder stages. Although the competing methods
also show an increase near the final decoder stage, SASFormer exhibits a
more pronounced late-decoder re-alignment and reaches the highest final
rain--noise feature similarity. This trajectory is consistent with the
proposed stage-aware strategy: degradation-dependent information remains
available in the encoder and early decoder, whereas later decoder stages
transform these features toward a common restoration-oriented
representation. This analysis characterizes representation dynamics;
restoration quality is independently evaluated using the standard
PSNR/SSIM results.

\section{Experiments}
\label{Experiments}

Following the experimental setups in prior works \cite{zamfir2025complexity, zeng2025vision}, we conduct two types of image restoration experiments: (i) All-in-One restoration and (ii) Composited degradation restoration. In the All-in-One setting, a single unified model is trained to handle multiple degradation types, specifically considering three and five distinct degradations. In the Composited degradation setting, the model is trained to handle both individual degradations and combinations of up to three different degradation types. Extensive experiments are conducted to evaluate SASFormer through comparisons with state-of-the-art methods, representation analyses, ablation studies, generalization
experiments, and efficiency evaluations.

\begin{figure*}[tb]
	\centering
	\includegraphics[width=1\linewidth]{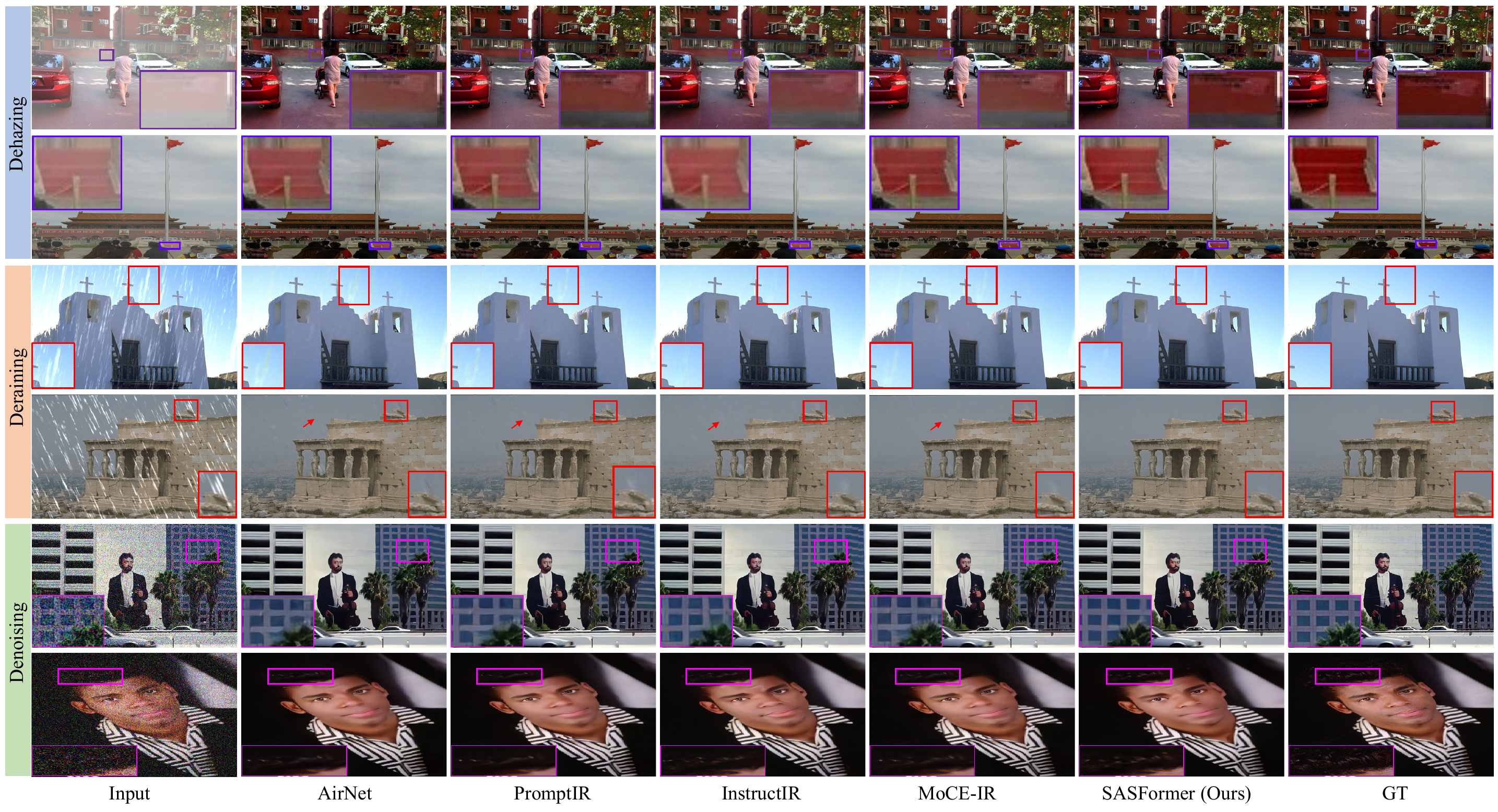} 
	\caption{Visual comparison of SASFormer and state-of-the-art methods on challenging cases under the all-in-one image restoration setting involving three degradation types. SASFormer effectively removes globally distributed haze and spatially localized rain streaks, while preserving image sharpness and fine details in detail-sensitive denoising tasks, consistently achieving high-quality restoration results.}
	\label{multi-task-visual-3D}
\end{figure*}

\begin{figure*}[tb]
	\centering
	\includegraphics[width=1\linewidth]{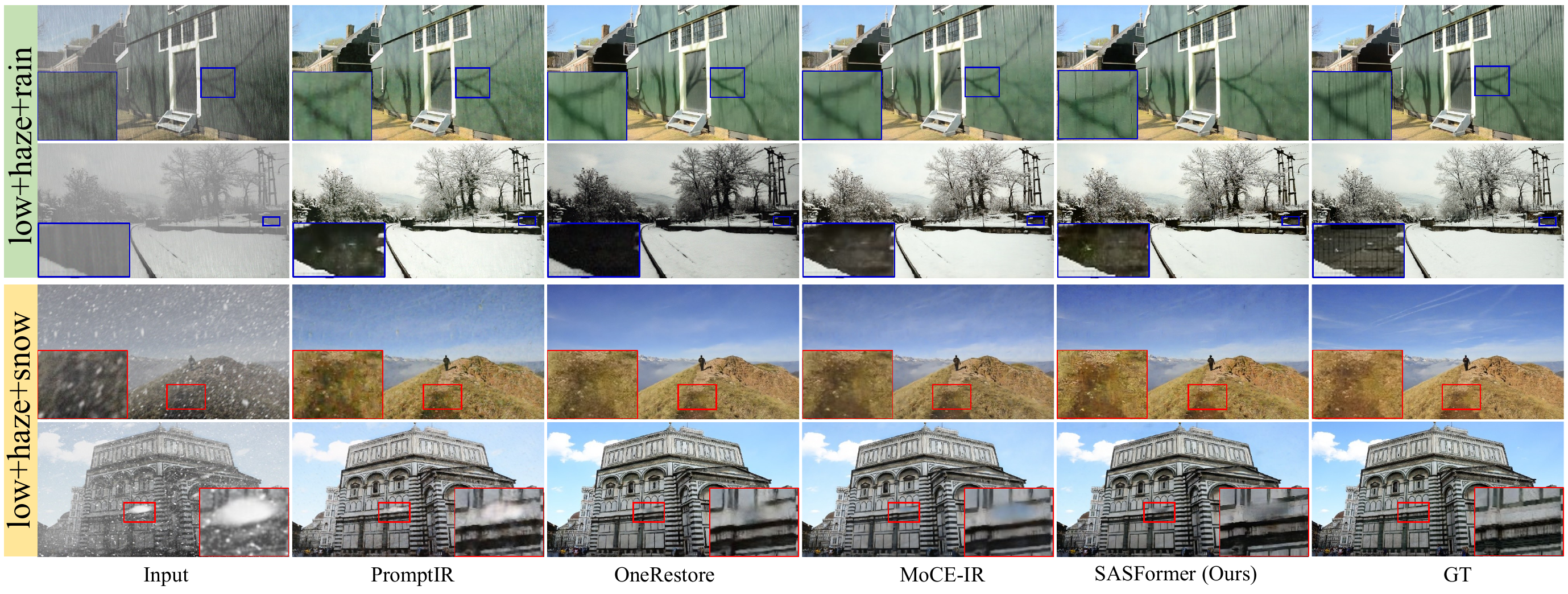} 
	\caption{Visual comparison of SASFormer against state-of-the-art methods across three challenging mixed degradation scenarios.}
	\label{multi-task-cdd11-3mix}
\end{figure*}

\subsection{Experimental Settings}
\textbf{Datasets.} For the all-in-one setting, following prior work~\cite{potlapalli2023promptir, all-in-one_IDR_zhang2023ingredient}, we use the following datasets: for image denoising, BSD400~\cite{arbelaez2010contour} and WED~\cite{ma2016waterloo} are combined, with Gaussian noise added at levels $\sigma \in \{15,25,50\}$, and CBSD68~\cite{martin2001database} is used for testing; for image deraining, Rain100L~\cite{yang2017deep} is employed; for dehazing, the SOTS~\cite{li2018benchmarking} dataset is used; and for deblurring and low-light enhancement, GoPro~\cite{nah2017deep} and LOL-v1~\cite{wei2018deep} are adopted, respectively. These datasets are combined under three- or five-degradation settings for training over 100 epochs, with direct evaluation conducted across the different tasks. In the composited degradation setting, the CDD11~\cite{guo2024onerestore} dataset is used, and the model is trained for 200 epochs.

\textbf{Implementation Details.} Our SASFormer provides an end-to-end trainable solution without the need for pretraining any individual modules. The network adopts a four-level encoder-decoder architecture, with each level containing a different number of Transformer blocks: [2, 3, 3, 4] from the first to the fourth level.
Experiments are conducted using PyTorch on a single NVIDIA GeForce RTX4090 GPU. The initial learning rate is set to $1e^{-4}$ and is gradually reduced following a cosine annealing schedule. During training, input images are cropped into $128\times128$ pixel patches with a batch size of 12.
The network is optimized using the L1 loss function and the Adam optimizer with parameters $\beta_1 = 0.9$ and $\beta_2 = 0.999$. Data augmentation is applied via random horizontal and vertical flips. 
For fair comparison, our training strategy—including batch composition, data augmentation, and dataset splits—follows standard practices established in prior works~\cite{potlapalli2023promptir,cui2025adair,zamfir2025complexity,guo2024onerestore} across the 3-, 5-, and 11-degradation settings.

\textbf{Temperature annealing.}
The channel-selection temperature $\tau^{\mathrm{ch}}$ and the
attention-selection temperature $\tau^{\mathrm{att}}$ are maintained as
separate variables for their respective continuous relaxations. Both
temperatures are updated once at the beginning of each training epoch and
remain fixed within that epoch. Given the current epoch index $t$ and the
total number of training epochs $T$, we use exponential annealing:
\begin{equation}
	\tau_t^{\mathrm{ch}}
	=
	\tau_t^{\mathrm{att}}
	=
	\tau_{\mathrm{start}}
	\left(
	\frac{\tau_{\mathrm{end}}}
	{\tau_{\mathrm{start}}}
	\right)^{t/T},
\end{equation}
where $t$ is the epoch index used by the training schedule and $T$ is the
total number of training epochs. This schedule yields
$\tau_0=\tau_{\mathrm{start}}$ at the beginning of training and
$\tau_T=\tau_{\mathrm{end}}$ at the end of training. We set
$\tau_{\mathrm{start}}=0.5$ and $\tau_{\mathrm{end}}=0.05$. Here,
$\tau^{\mathrm{ch}}$ controls the Gumbel--Softmax relaxation for channel
selection, whereas $\tau^{\mathrm{att}}$ controls the rank-based Sigmoid
surrogate for attention selection.

\subsection{Comparison With State-of-the-Art Methods}

\textbf{Three-Degradation Setting.}
We compare the proposed SASFormer with both general image restoration methods and all-in-one image restoration methods. The general restoration methods include Restormer \cite{IM_zamir2022restormer}, NAFNet \cite{IM_NAFNetchen2022simple}, FSNet \cite{cui2023image}, and MambaIR \cite{guo2025mambair}. The all-in-one methods consist of blind restoration models, including AirNet \cite{all-in-one_AirNet_li2022all}, PromptIR \cite{potlapalli2023promptir}, AdaIR \cite{cui2025adair}, and MoCE-IR \cite{zamfir2025complexity}, as well as non-blind restoration models, including InstructIR \cite{conde2025instructir}, DA-RCOT \cite{10970105}, VLU-Net \cite{zeng2025vision}, Perceive-IR~\cite{Perceive-IR-24}, ClearAIR*~\cite{zhang2026clearair}, and R2R~\cite{wang2026retrieve}. All methods are jointly trained on three degradation types, namely haze, rain streaks, and Gaussian noise.
As reported in Tab.\ref{all-in-one-3T}, SASFormer achieves competitive restoration performance, with an average PSNR \textcolor{red}{0.10} dB higher than that of MEASNet\cite{yu2024multi}. Moreover, despite employing fewer parameters than other leading methods, 

SASFormer achieves competitive restoration performance across different tasks while maintaining an excellent balance between effectiveness and efficiency. Qualitative comparisons are presented in Fig.~\ref{multi-task-visual-3D}, where our method produces visually favorable results with improved detail recovery, more effective rain streak removal, and clearer restoration under severe haze degradation compared with competing approaches.

\noindent \textbf{Five-Degradation Setting.} 
Building upon the experimental setup in \cite{all-in-one_IDR_zhang2023ingredient, zamfir2025complexity}, we extend the three-degradation setting to include deblurring and low-light enhancement, providing a more comprehensive validation of our method's effectiveness. As shown in Tab.~\ref{all-in-one-5T}, our approach demonstrates stronger representational capability for complex degradations and effectively mitigates interference among multiple degradation types. Compared with MEASNet \cite{yu2024multi}, our model achieves an average PSNR improvement of \textcolor{red}{0.30} dB while reducing parameters by \textcolor{red}{24.1\%}, highlighting its superior efficiency and performance.

\noindent \textbf{Composited Degradations.}
To further validate the robustness of our method in more complex and mixed degradation scenarios, and following standard practices in the field, we conduct experiments on the CDD-11 dataset \cite{guo2024onerestore}. This dataset includes various single degradations such as rain, haze, snow, and low-light conditions, as well as multiple composited degradations, resulting in a total of 11 distinct restoration settings.
As shown in Tab.~\ref{cdd-11-tab_all-in-one}, our method achieves average PSNR gains of \textcolor{red}{0.74} dB and \textcolor{red}{1.01} dB over R2R \cite{wang2026retrieve} and MoCE-IR-S \cite{zamfir2025complexity}, respectively. Fig.~\ref{multi-task-cdd11-3mix} further presents visual comparisons on this dataset, where it can be observed that our method produces superior results in both fine-grained texture restoration and global structural consistency. These results further demonstrate the effectiveness and superiority of the proposed approach.

\noindent \textbf{Generalization Performance Evaluation.}
To evaluate the generalization capability of our model, we tested it on three datasets not seen during training: Rain100H \cite{yang2017deep} for heavy deraining, HIDE \cite{shen2019human} for human-centric deblurring, and LoLv2-Real \cite{yang2021sparse} for low-light image enhancement. Tab.~\ref{tab-generalization} reports the results of models trained under five degradation conditions directly applied to these test sets. While MoCE-IR~\cite{zamfir2025complexity} benefits from Sobel filtering, which provides an advantage in texture extraction and leads to slightly better deblurring performance, our method demonstrates clear improvements in heavy deraining, real low-light enhancement, and average PSNR. These results further validate the effectiveness of SASFormer in feature representation and its strong generalization ability.

\noindent \textbf{Zero-Shot Generalization}
Following the experimental settings of MoCE-IR \cite{zamfir2025complexity}, we evaluate representative blind all-in-one image restoration methods, including AirNet \cite{all-in-one_AirNet_li2022all}, PromptIR \cite{potlapalli2023promptir}, and MoCE-IR \cite{zamfir2025complexity}, in terms of their zero-shot generalization ability. All methods are trained with three degradation types and directly evaluated on unseen degradations without any additional fine-tuning. For a fair comparison, we adopt the official pre-trained checkpoints of AirNet, PromptIR, and MoCE-IR, while our model also uses the corresponding checkpoint trained with the same three-degradation setting.
Specifically, we consider several deblurring tasks, including defocus deblurring on DPDD \cite{ai2024multimodal}, as well as motion deblurring on GoPro \cite{nah2017deep} and HIDE \cite{shen2019human}, where these degradation types are not included during training. PSNR, SSIM, and LPIPS are employed as evaluation metrics.
As shown in Tab.~\ref{tab-zeroshot}, our method achieves the best PSNR performance on all three datasets while maintaining competitive SSIM and LPIPS results. These results demonstrate the strong zero-shot generalization capability of our approach to unseen degradation types.

\begin{figure}[tb]
	\centering
	\includegraphics[width=1\linewidth]{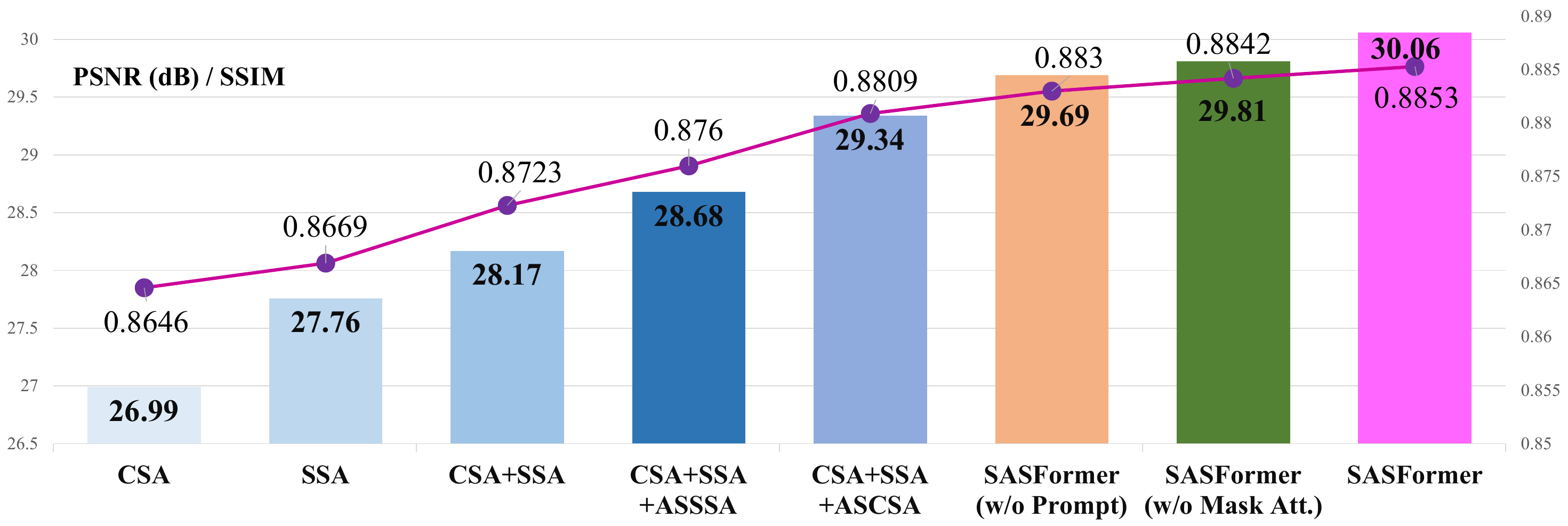} 
	\caption{Ablation study validating the contributions of key components. The results of average PSNR across \textit{composited degradation} scenarios are presented to validate the contribution of each component to the overall performance.}
	\label{ablation-key-com}
\end{figure}

\subsection{Ablation Analysis}
We perform a detailed analysis of each component in our method to validate its effectiveness. 
We primarily conduct ablation studies under the \textit{Three Degradations} and \textit{Composited Degradations} evaluation settings.

\noindent \textbf{Key Components.}
As shown in Fig.~\ref{ablation-key-com}, we present the ablation results of key components under the \textit{Composited Degradations} setting to evaluate their contributions to SASFormer’s performance. Replacing the decoder’s CSA+SSA with ASSSA and ASCSA yields PSNR gains of 0.51 dB and 1.17 dB, respectively, validating the effectiveness of the adaptive selection mechanism. Further incorporating implicit prompts and AASM enhances the model’s degradation awareness and feature representation, providing additional PSNR improvements of 0.23 dB and 0.13 dB, and resulting in the best average performance.

\begin{figure}[tb]
	\centering
	\includegraphics[width=1\linewidth]{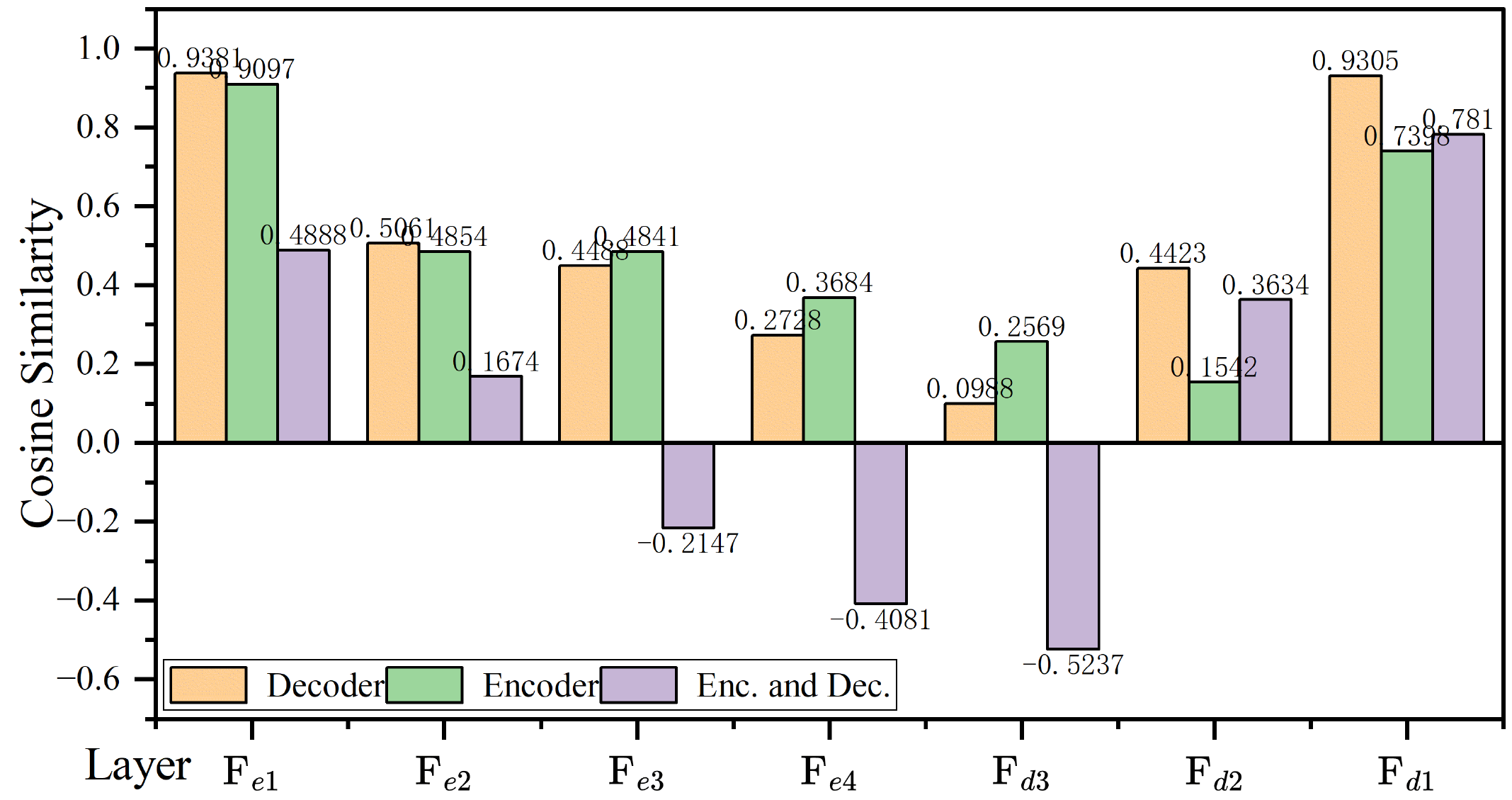} 
	\caption{Layer-wise cross-degradation cosine similarity under different stage-wise configurations of ASCSA and ASSSA, supporting the proposed stage-aware design.}
	\label{ablation-cosine-Configuration}
\end{figure}

\begin{table}[htb]
	\centering
	\caption{Ablation study for validating the stage-aware design with different configurations of ASCSA and ASSSA. All models are trained and evaluated under the \textit{three degradation} setting.}
	\label{Configuration_ablation}
	\begin{tabular}{c|ccc}
		\hline
		Configuration & Encoder & Encoder + Decoder & Decoder \\
		\hline
		Params. & 23.61M & 24.57M & 23.69M\\
		PSNR/SSIM & 32.74/0.918 & 32.83/0.919 & 32.95/0.920 \\
		\hline
	\end{tabular}
\end{table}

\begin{figure}[t]
	\centering
	
	\includegraphics[width=1\linewidth]{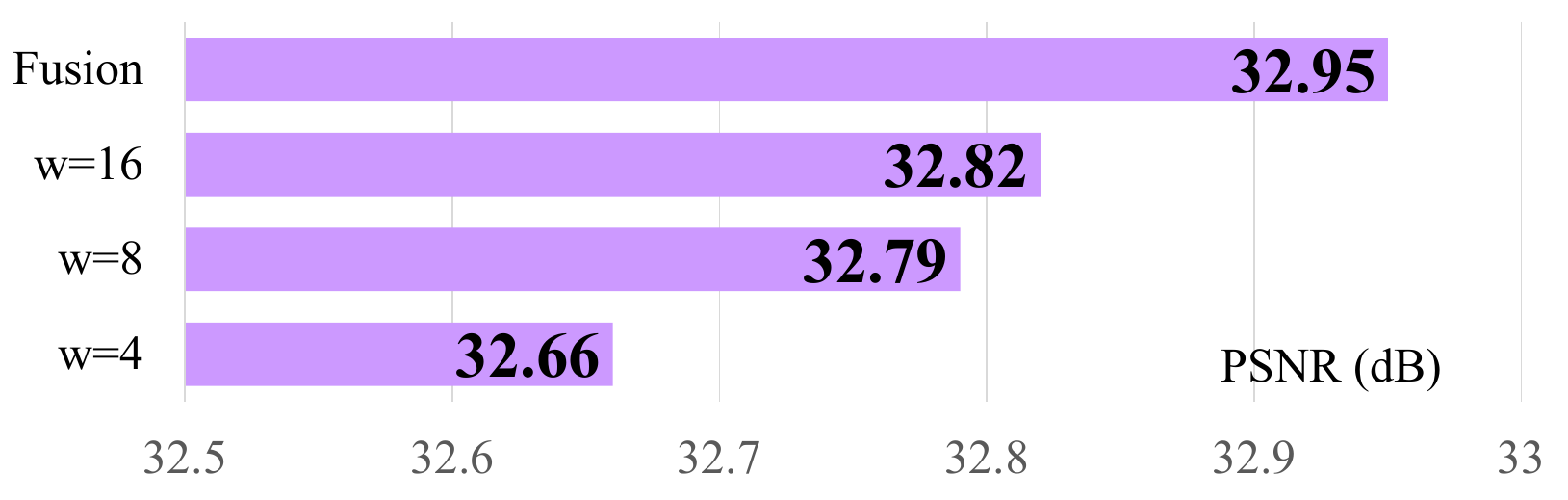}
	\caption{Ablation results on spatial window size $ w $ under \textit{three degradation} settings.}
	\label{ablation-space}
\end{figure}

\noindent\textbf{Validation of the Stage-Aware Design.}
To validate the proposed stage-aware design, we compare three configurations of ASCSA and ASSSA: encoder-only, encoder--decoder, and decoder-only. As reported in Tab.~\ref{Configuration_ablation}, they achieve average PSNR values of 32.74 dB, 32.83 dB, and 32.95 dB, respectively, on the three-degradation benchmark. The decoder-only configuration performs best, indicating that extending adaptive selection to more stages does not necessarily improve restoration quality.
Fig.~\ref{ablation-cosine-Configuration} further shows that applying adaptive selection throughout both encoding and decoding leads to stronger cross-degradation feature divergence at intermediate stages. Together with its lower PSNR, this suggests that excessive early-stage selection may interfere with the representation preservation required for subsequent reconstruction. In contrast, decoder-side selection enables adaptive feature refinement during reconstruction and achieves the best performance among the evaluated configurations. These results are consistent with our stage-wise analysis and provide empirical support for the proposed stage-aware design.

\begin{figure}[tb]
	\centering
	\includegraphics[width=1\linewidth]{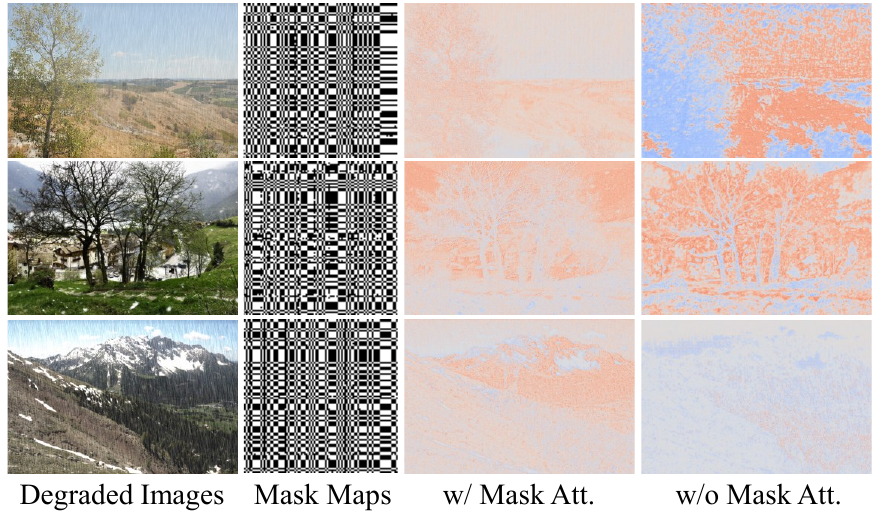} 
	\caption{Visualization of mask attention features. The masks vary across degradations and emphasize intrinsic image content. }
	\label{visual-mask}
\end{figure}

\noindent \textbf{Spatial Selectivity.}
Fig.~\ref{ablation-space} presents the ablation results for different spatial attention window sizes. Small windows overemphasize local regions and amplify degradation cues, reducing performance. Larger windows capture global structure and improve performance, but too large loses fine details. Multi-scale windows balance local details and global structure, achieving optimal performance.

\begin{figure}[tb]
	\centering
	\includegraphics[width=1\linewidth]{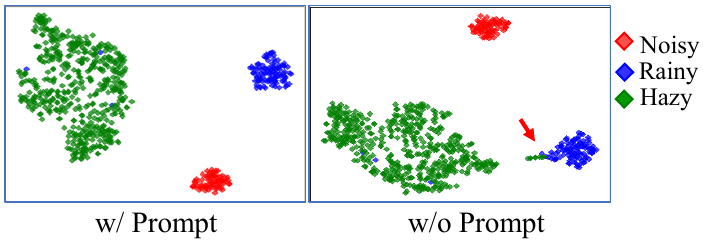} 
	\caption{t-SNE visualization with and without implicit prompts, showing improved multi-degradation feature discrimination with prompting.}
	\label{tsne-ablation-prompt}
\end{figure}

\noindent \textbf{AASM and Prompt.} 
As shown in Fig.~\ref{visual-mask}, the channel-wise mask attention highlights that different degradations produce distinct activation patterns, with features increasingly focusing on intrinsic image content while suppressing degradation cues. This demonstrates the effectiveness of our adaptive attention selection in enhancing feature robustness. 
As shown in Fig.~\ref{tsne-ablation-prompt}, the t-SNE visualizations with and without prompt guidance reveal that incorporating prompts leads to clearer separation and more compact clustering of degradation features. This indicates that the model is capable of learning robust multi-degradation feature representations even without prompts, while the prompt mechanism further enhances feature discriminability and separability.

\noindent \textbf{Effect of the number of prompts.}
We further study the effect of the number of prompts under the composited degradation setting, as reported in Tab.~\ref{tab:prompt_ablation}. Introducing prompts consistently improves the restoration performance over the prompt-free baseline, indicating that prompt-based modulation is beneficial for degradation-adaptive restoration. The PSNR increases from 29.6924 dB without prompts to 30.0593 dB with three prompts, while further increasing the number of prompts brings only marginal gains. Therefore, we adopt three prompts as a trade-off between restoration accuracy and model complexity.

\begin{figure}[tb]
	\centering
	\includegraphics[width=1\linewidth]{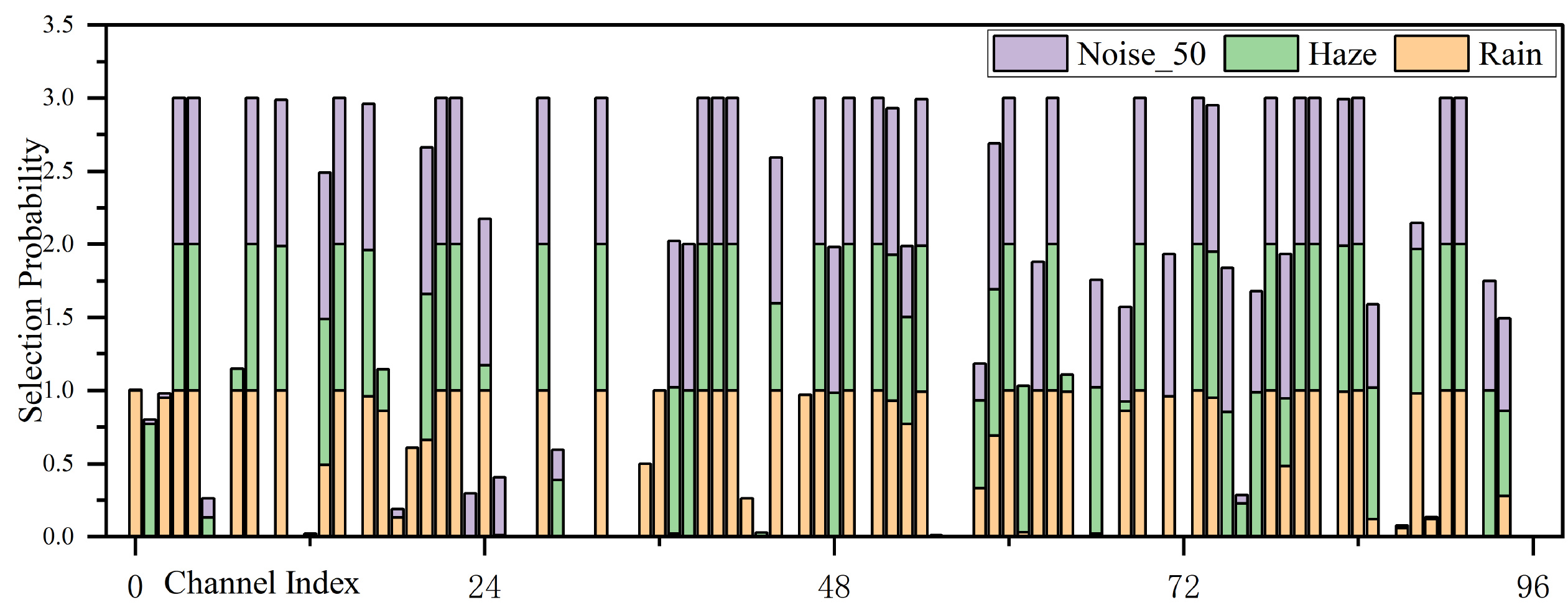} 
	\caption{Channel selection probabilities of ASCSA under \textit{three degradation} types. The horizontal axis denotes the channel index, and the vertical axis denotes the selection probability.}
	\label{visual-selection}
\end{figure}

\begin{figure}[tb]
	\includegraphics[width=1\linewidth]{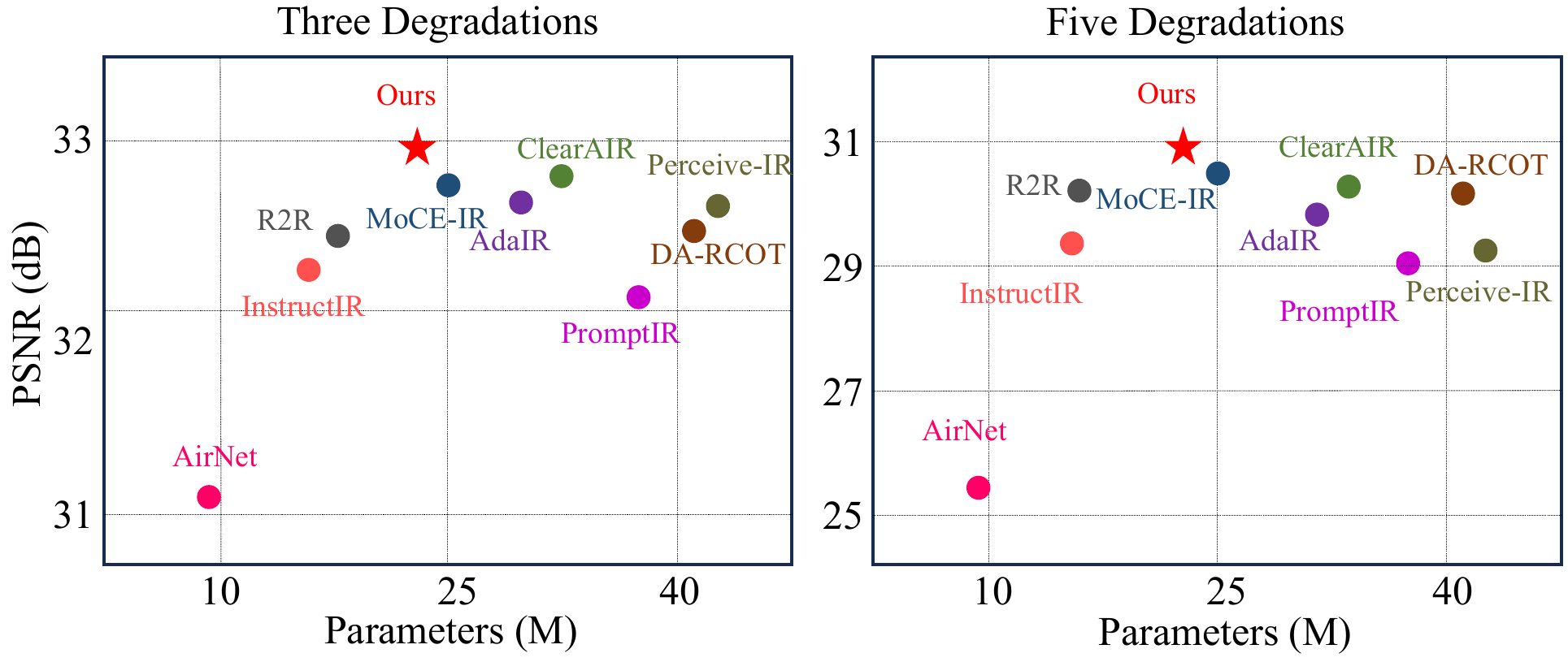}
	\caption{Model parameter versus PSNR comparison under three- and five-degradation settings. Our SASFormer achieves a favorable performance-efficiency trade-off, delivering superior restoration performance with a compact parameter count among the compared methods.}
	\label{visual-Parameter}
\end{figure}

\begin{table}[tb]
	\centering
	\caption{Effect of the number of prompts on image restoration performance under the \textit{composited degradation} setting.}
	\label{tab:prompt_ablation}
	\begin{tabular}{c|ccc}
		\hline
		Number of Prompts & Params. & PSNR (dB) & SSIM \\
		\hline
		0 & 23.67M & 29.6924 & 0.8832  \\
		1 & 23.68M & 29.7912 & 0.8842  \\
		2 & 23.68M & 29.9105 & 0.8853  \\
		3 & 23.69M & 30.0593  & 0.8853  \\
		4 & 23.69M & 30.0598 & 0.8854  \\
		5 & 23.69M & 30.0612 & 0.8854  \\
		\hline
	\end{tabular}
\end{table}

\begin{table}[tb] 
	\caption{Model parameters and computational complexity are evaluated under the \textit{three degradations} setting. GMACS are computed on an input image of size 256 $\times $ 256 using an NVIDIA RTX4090 GPU.}
	\label{tab:Parameters}
	\centering
	\setlength{\tabcolsep}{9pt} 
	\renewcommand{\arraystretch}{1.1}  
	\begin{tabular}{c|ccc} 
		\hline  
		Method & Params. & GMACS  & PSNR/SSIM \\ 
		\hline
		PromptIR & 35.59M & 158.40 & 32.06/0.913\\ 
		AdaIR & 28.77M & 147.45 & 32.69/0.918\\
		MoCE-IR & 25.35M & 87.31 & 32.73/0.917\\
		SASFormer (Ours) & 23.69M & 125.47 & 32.95/0.920\\ 
		\hline	
	\end{tabular}				
\end{table}

\begin{table}[t] 
	\caption{Memory usage and inference time under the \textit{five degradations} test sets. Peak GPU memory usage and inference time at a resolution of 256$\times$256 (NVIDIA RTX4090) are also reported.}
	\label{tab:Runtime}
	\centering
	\setlength{\tabcolsep}{8pt} 
	\renewcommand{\arraystretch}{1.1}  
	\begin{tabular}{c|ccc} 
		\hline  
		Method & Memory & Runtime & PSNR/SSIM \\ 
		\hline
		PromptIR & 1113.40M & 92.15±0.56ms & 29.15/0.904\\ 
		AdaIR & 1318.00M & 97.12±0.82ms & 30.20/0.910\\
		MoCE-IR & 336.79M & 80.84±0.94ms & 30.58/0.919\\
		SASFormer (Ours) & 1253.46M & 90.85±0.43ms & 30.94/0.918\\ 
		\hline	
	\end{tabular}				
\end{table}

\noindent \textbf{Analysis of adaptive channel selection.}
Fig.~\ref{visual-selection} visualizes the channel selection probabilities of ASCSA under three degradation types. Samples with the same degradation tend to activate similar channel groups, while some channels are selectively emphasized for specific degradations. This indicates that ASCSA learns degradation-adaptive channel allocation instead of using a fixed channel attention pattern, thereby improving its ability to model diverse degradation characteristics.

\noindent \textbf{Complexity-Performance Trade-off.}
Our method achieves a favorable trade-off between model complexity and restoration performance by introducing a stage-aware selective self-attention mechanism. As shown in Fig.~\ref{visual-Parameter}, with an input size of $256 \times 256$, the model contains only 23.69M parameters with moderate FLOPs. Compared with previous state-of-the-art methods, it not only significantly improves restoration quality but also maintains a compact model size, demonstrating an excellent balance between performance and efficiency.

\noindent \textbf{Parameter Efficiency, Computational Cost, Memory Usage, and Runtime.}
As shown in Tab.~\ref{tab:Parameters}, SASFormer has the fewest parameters among the compared methods, demonstrating its parameter efficiency. Computational cost, however, follows a different pattern: SASFormer requires fewer GMACS than PromptIR~\cite{potlapalli2023promptir} and AdaIR~\cite{cui2025adair}, but more than MoCE-IR~\cite{zamfir2025complexity} because of its multi-scale spatial attention.
Memory usage and inference runtime are evaluated separately from parameter count and GMACS. SASFormer uses more memory than MoCE-IR and PromptIR but less than AdaIR under the evaluated settings. The inference-time comparisons in Tab.~\ref{tab:Runtime} similarly show that SASFormer is slower overall than MoCE-IR but faster overall than PromptIR and AdaIR. These results reflect the additional computational and memory costs of multi-scale spatial attention despite the compact parameter size.
This trade-off is accompanied by average PSNR gains of 0.22, 0.36, and 1.01 dB over MoCE-IR on the 3-, 5-, and 11-degradation test sets, respectively. Thus, SASFormer offers improved average restoration accuracy with fewer parameters, at the expense of higher computational cost, memory usage, and inference time relative to MoCE-IR.

\section{Conclusion}
\label{sec:conclusion}

In this paper, we proposed SASFormer, a stage-aware selective attention transformer for all-in-one blind image restoration. We identify distinct representation requirements across encoding and decoding under diverse degradations, and accordingly introduce a stage-aware architecture that differentiates feature modeling between the two stages. SASFormer separates encoder-side representation preservation from decoder-side selective reconstruction. Specifically, the encoder uses established hybrid channel-spatial attention to build diverse degradation-aware and content-preserving representations, while the decoder employs adaptive selective channel and spatial attention with implicit prompt-guided modulation to suppress degradation-specific responses and recover intrinsic image structures. Extensive experiments on 3-, 5-, and 11-degradation settings demonstrate that SASFormer achieves strong restoration accuracy with favorable parameter efficiency. Stage-wise representation analysis and ablation studies provide empirical support for the proposed stage-aware design, with decoder-side selection achieving the best restoration performance among the evaluated configurations. Nevertheless, the use of spatial attention slightly increases memory consumption, particularly for high-resolution restoration. In future work, we will investigate more efficient and lightweight selective attention mechanisms guided by stage-specific functions, and extend them to open-set real-world degradations and video restoration tasks.


\vspace{11pt}

\vspace{-33pt}
\begin{IEEEbiography}[{\includegraphics[width=1in,height=1.25in,clip,keepaspectratio]{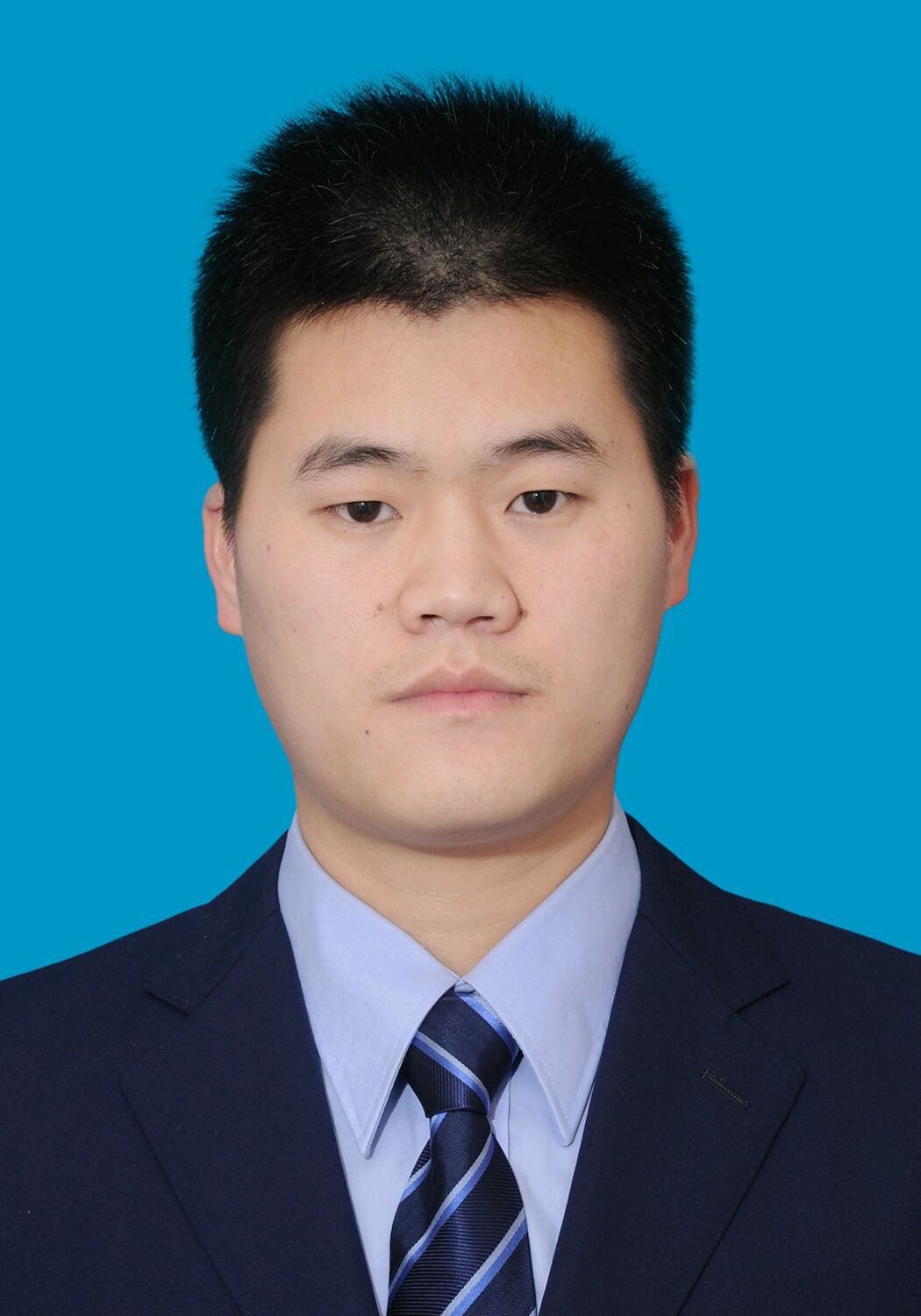}}]{Xiangpeng Tian} received his B.S. degree in Electronic Information Science and Technology from Hubei Normal University, China, in 2010, and his M.S. degree in Signal and Information Processing from Zhejiang Sci-Tech University, China, in 2013. He is currently pursuing a Ph.D. degree in Electronic Information at Sichuan University, China.
	His research interests include image restoration, and high-level vision. He has published papers on image restoration in journals and conferences, including CVPR.
\end{IEEEbiography}

\vspace{-33pt}
\begin{IEEEbiography}[{\includegraphics[width=1in,height=1.25in,clip,keepaspectratio]{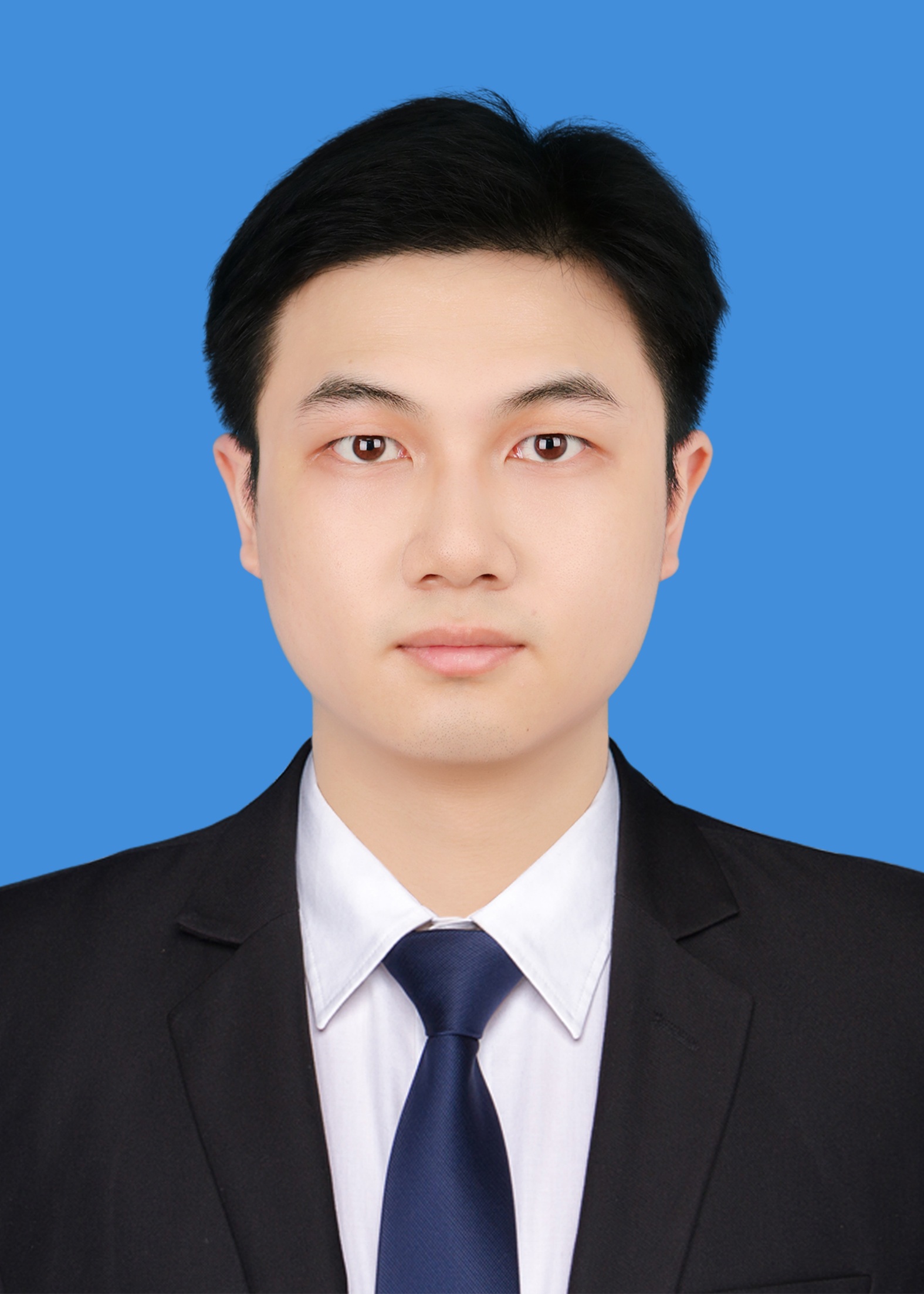}}]{Mingyang Li} received the B.S. degree from Sichuan University, Chengdu, China, in 2024. He is currently pursuing the M.S. degree at the College of Information and Communication Engineering, Sichuan University, Chengdu, China. His research interests include image super-resolution and image restoration.
\end{IEEEbiography}

\vspace{-33pt}
\begin{IEEEbiography}[{\includegraphics[width=1in,height=1.25in,clip,keepaspectratio]{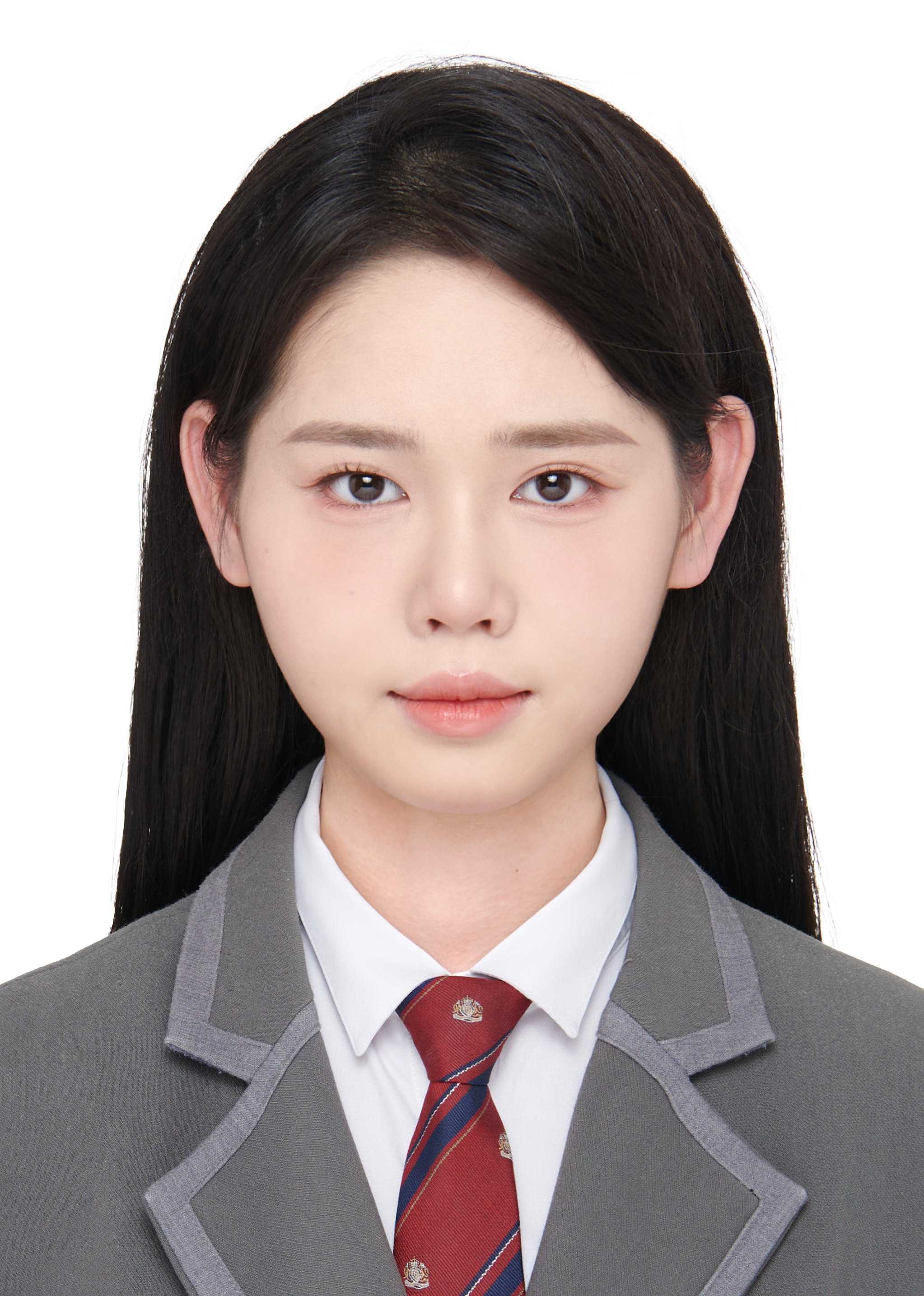}}]{Fajing Liu} received her B.S. degree from Tiangong University, China, in 2024. She is currently pursuing her M.S. degree in the College of Electronics and Information Engineering at Sichuan University, China. Her research interests include computer vision, particularly image restoration, object detection, and related topics.
\end{IEEEbiography}

\vspace{-33pt}
\begin{IEEEbiography}[{\includegraphics[width=1in,height=1.25in,clip,keepaspectratio]{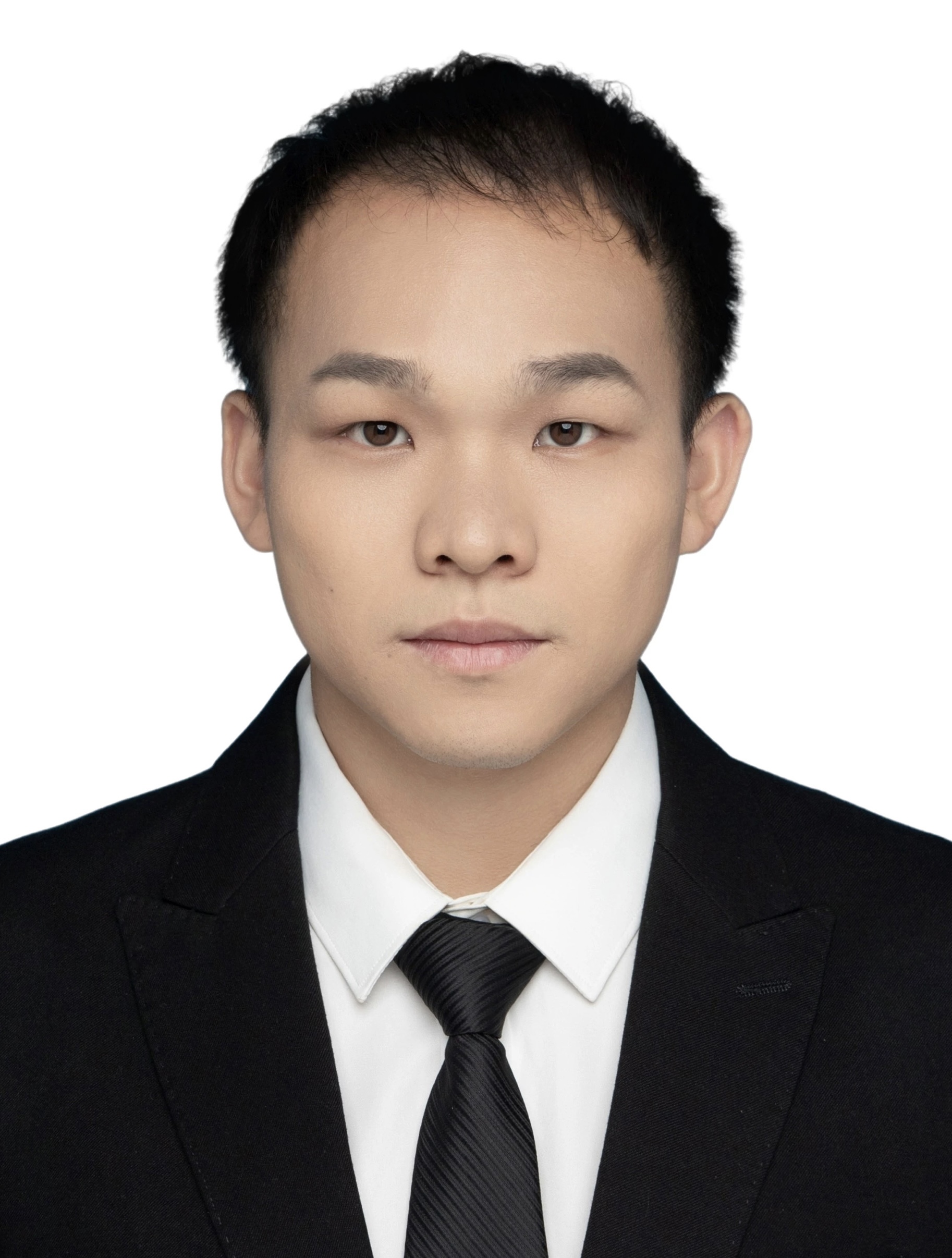}}]{Xiaofeng Wang} received the B.S. degree from Chongqing University of Technology, Chongqing, China, in 2021, and the M.S. degree in electronic science and technology from the College of Electronics and Information Engineering, Sichuan University, Chengdu, China, in 2024. He is currently pursuing the Ph.D. degree in electronic information with the College of Electronics and Information Engineering, Sichuan University. His research interests include image restoration, object detection, and their integration for robust visual perception.
\end{IEEEbiography}

\vspace{-33pt}
\begin{IEEEbiography}[{\includegraphics[width=1in,height=1.25in,clip,keepaspectratio]{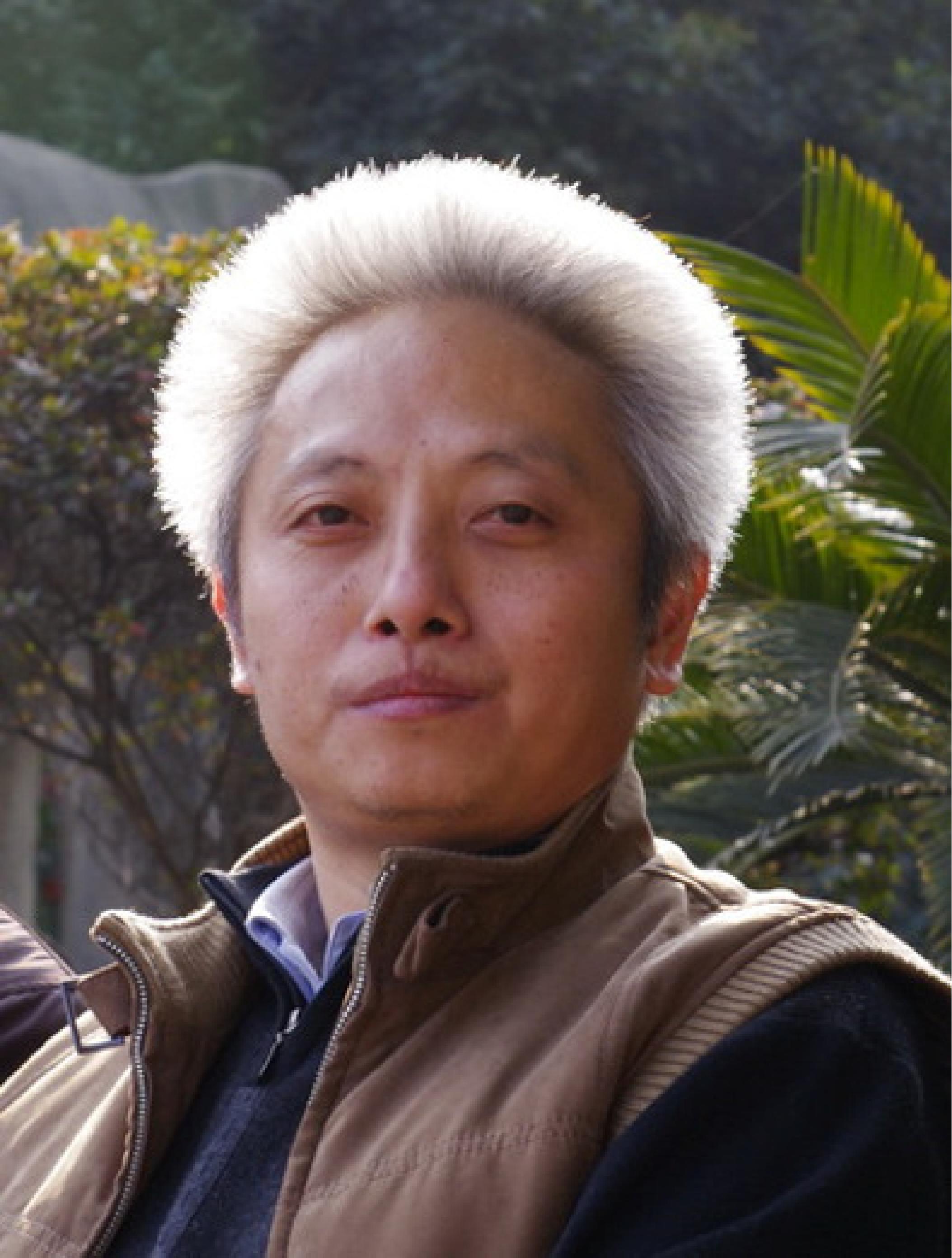}}]{Xiaohai He}(Member, IEEE) received the B.S., M.S., and Ph.D. degrees from Sichuan University, Chengdu, China, in 1985, 1991, and 2002, respectively.	
	He is currently a Professor with the College of Electronics and Information Engineering, Sichuan University. His current research interests include image processing, pattern recognition, computer vision, image compression, and software engineering. 
	Prof. He is an Editor of the Journal of Information and Electronic Engineering and was an Editor of the Journal of Data Acquisition and Processing. He is a Senior Member of the Chinese Institute of Electronics.
\end{IEEEbiography}

\vspace{-33pt}
\begin{IEEEbiography}[{\includegraphics[width=1in,height=1.25in,clip,keepaspectratio]{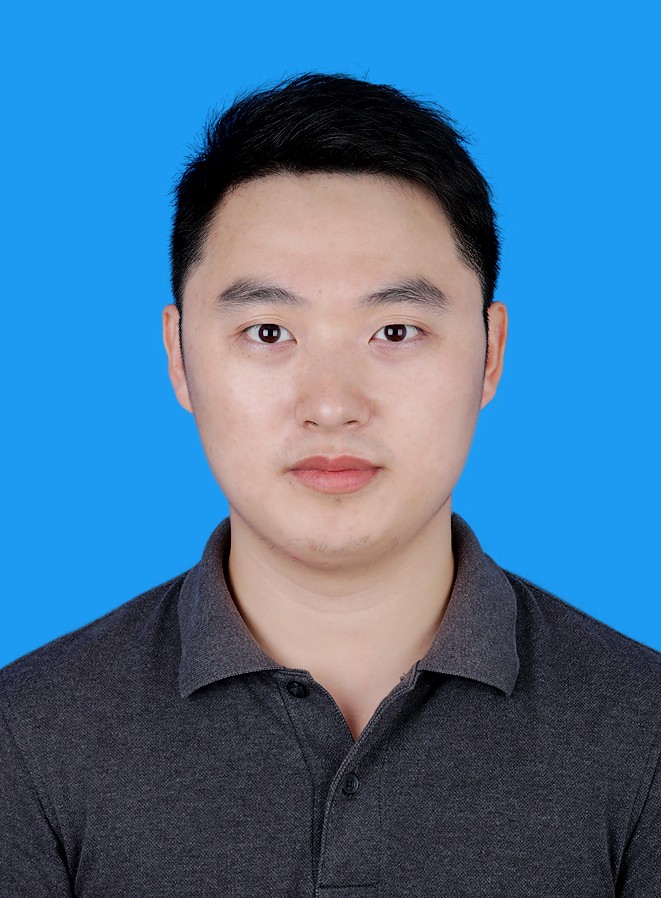}}]{Chao Ren}(Member, IEEE) received the B.S.degree in electronics and information engineering and Ph.D degree in communication and information system from Sichuan University, Chengdu, China, in 2012 and 2017, respectively. From 2015 to 2016, he was a Visiting Scholar with the Department of Electrical and Computer Engineering, University of California at San Diego, La Jolla, CA, USA. He is currently an Associate Research Professor at Sichuan University College of Electronics and Information Engineering. His research interests include inverse problems in image and video processing. He has authored more than 70 papers in journals/conferences such as TPAMI, TIP, CVPR, ICCV, ECCV, NeurIPS, etc. He has won the Huawei Spark Award and was selected for the National Post-Doctoral Program for Innovative Talents of China.
\end{IEEEbiography}

\end{document}